\documentclass{article}
\pdfoutput=1

 \usepackage[preprint]{neurips_2019}

\usepackage[utf8]{inputenc} 
\usepackage[T1]{fontenc}    
\usepackage{hyperref}       
\usepackage{url}            
\usepackage{booktabs}       
\usepackage{amsfonts}       
\usepackage{nicefrac}       
\usepackage{microtype}      

\usepackage{subfig}
\usepackage{graphicx}
\usepackage{algorithm2e}
\usepackage{amsmath,amsthm,amssymb}
\usepackage{xifthen}
\usepackage{multirow}
\usepackage{comment}

\allowdisplaybreaks

\makeatletter
\newcommand\footnoteref[1]{\protected@xdef\@thefnmark{\ref{#1}}\@footnotemark}
\makeatother

\usepackage{array}
\newcolumntype{L}[1]{>{\raggedright\let\newline\\\arraybackslash\hspace{0pt}}m{#1}}
\newcolumntype{C}[1]{>{\centering\let\newline\\\arraybackslash\hspace{0pt}}m{#1}}
\newcolumntype{R}[1]{>{\raggedleft\let\newline\\\arraybackslash\hspace{0pt}}m{#1}}

\mathchardef\mhyphen="2D
\newcommand{\llca}{\ell^{lca}}
\newcommand{\Wb}{\mathbb{W}}
\newcommand{\Pb}{\mathbf{P}}
\newcommand{\Eb}{\mathbf{E}}

\def\tRhelp#1#2\relax{L_{\csname dom#1\endcsname#2}} 

\def\eRhelp#1#2\relax{\hat{L}_{\csname set#1\endcsname#2}}
\newcommand{\loss}[1]{l\ifthenelse{\isempty{#1}{}}{}{\left(#1\right)}}

\newcommand{\normal}[2]{\mathcal{N}\left(#1,#2\right)}

\DeclareMathOperator*{\argmin}{arg\,min}
\DeclareMathOperator*{\argmax}{arg\,max}

\newcommand{\mydefv}[1]{\expandafter\newcommand\csname v#1\endcsname{\mathbf{#1}}}
\newcommand{\mydefallv}[1]{\ifx#1\mydefallv\else\mydefv{#1}\expandafter\mydefallv\fi}
\mydefallv abekmsuvwxyz\mydefallv 

\newcommand{\mydefvsym}[1]{\expandafter\newcommand\csname v#1\endcsname{\boldsymbol{\csname #1\endcsname}}}
\newcommand{\mydefallvsym}[1]{\ifx#1\mydefallvsym\else\mydefvsym{#1}\expandafter\mydefallvsym\fi}
\mydefallvsym {sigma}{alpha}{gamma}{mu}\mydefallvsym 

\newcommand{\mydefm}[1]{\expandafter\newcommand\csname m#1\endcsname{\mathbf{#1}}}
\newcommand{\mydefallm}[1]{\ifx#1\mydefallm\else\mydefm{#1}\expandafter\mydefallm\fi}
\mydefallm ABCIKLMOSTVWXZ\mydefallm 

\newcommand{\mydefmsym}[1]{\expandafter\newcommand\csname m#1\endcsname{\boldsymbol{\csname #1\endcsname}}}
\newcommand{\mydefallmsym}[1]{\ifx#1\mydefallmsym\else\mydefmsym{#1}\expandafter\mydefallmsym\fi}
\mydefallmsym {Sigma}{Gamma}{Phi}{gamma}\mydefallmsym 

\newcommand{\mydefalg}[1]{\expandafter\newcommand\csname alg#1\endcsname{\mathcal{#1}}}
\newcommand{\mydefallalg}[1]{\ifx#1\mydefallalg\else\mydefalg{#1}\expandafter\mydefallalg\fi}
\mydefallalg A\mydefallalg 

\newcommand{\mydefdom}[1]{\expandafter\newcommand\csname dom#1\endcsname{\mathcal{#1}}}
\newcommand{\mydefalldom}[1]{\ifx#1\mydefalldom\else\mydefdom{#1}\expandafter\mydefalldom\fi}
\mydefalldom HSTVX\mydefalldom 

\newcommand{\mydefset}[1]{\expandafter\newcommand\csname set#1\endcsname{\mathcal{#1}}}
\newcommand{\mydefallset}[1]{\ifx#1\mydefallset\else\mydefset{#1}\expandafter\mydefallset\fi}
\mydefallset BCGHPQRSTVX\mydefallset 

\newcommand{\mydefdistr}[1]{\expandafter\newcommand\csname distr#1\endcsname{\mathcal{D}_{\csname dom#1\endcsname}}}
\newcommand{\mydefalldistr}[1]{\ifx#1\mydefalldistr\else\mydefdistr{#1}\expandafter\mydefalldistr\fi}
\mydefalldistr DHSTVX\mydefalldistr 

\newcommand{\mydefspace}[1]{\expandafter\newcommand\csname space#1\endcsname{\mathcal{#1}}}
\newcommand{\mydefallspace}[1]{\ifx#1\mydefallspace\else\mydefspace{#1}\expandafter\mydefallspace\fi}
\mydefallspace DFGHKLMPRUVXYZ\mydefallspace 

\newcommand{\mydeff}[1]{\expandafter\newcommand\csname f#1\endcsname[2][]{#1##1\ifthenelse{\equal{##2}{}}{}{\!\left(##2\right)}}}
\newcommand{\mydefallf}[1]{\ifx#1\mydefallf\else\mydeff{#1}\expandafter\mydefallf\fi}
\mydefallf cdfghkwCFGHMRT{PV}\mydefallf 

\newcommand{\mydeffsym}[1]{\expandafter\newcommand\csname f#1\endcsname[2][]{\csname #1\endcsname##1\ifthenelse{\equal{##2}{}}{}{\!\left(##2\right)}}}
\newcommand{\mydefallfsym}[1]{\ifx#1\mydefallfsym\else\mydeffsym{#1}\expandafter\mydefallfsym\fi}
\mydefallfsym {phi}{epsilon}{eta}\mydefallfsym 

\newcommand{\mydefnset}[1]{\expandafter\newcommand\csname nset#1\endcsname{\mathbb{#1}}}
\newcommand{\mydefallnset}[1]{\ifx#1\mydefallnset\else\mydefnset{#1}\expandafter\mydefallnset\fi}
\mydefallnset CNRSZ\mydefallnset 

\newcommand{\bigO}[1]{\mathcal{O}\left( #1 \right)}
\newcommand{\bigOmega}[1]{\Omega\left( #1 \right)}

\newtheorem{myth}{Theorem}
\newtheorem*{myth*}{Theorem}

\newtheorem*{mylem*}{Lemma}

\newtheorem{myex}{Example}
\newtheorem*{myex*}{Example}
\newtheorem{myprop}{Proposition}

\theoremstyle{definition}

\newtheorem*{myrem*}{Remark}

\newenvironment{reth}[1]
    {
\begingroup
\def\themyth{#1}
\begin{myth}
    }
    { 
\end{myth}
\addtocounter{myth}{-1}
\endgroup
    }

\newenvironment{reprop}[1]
    {
\begingroup
\def\themyprop{#1}
\begin{myprop}
    }
    { 
\end{myprop}
\addtocounter{myprop}{-1}
\endgroup
    }

\makeatletter
\def\th@plain{%
  \thm@notefont{}
  \itshape 
}
\def\th@definition{%
  \thm@notefont{}
  \normalfont 
}
\makeatother

\title{Foundations of Comparison-Based Hierarchical Clustering}

\author{%
Debarghya Ghoshdastidar \\
University of T\"{u}bingen, Department of Computer Science, \\
T\"{u}bingen, Germany \\
\texttt{debarghya.ghoshdastidar@uni-tuebingen.de} \\
\And
Micha\"el Perrot \\
Max-Planck-Institute for Intelligent Systems, \\
T\"{u}bingen, Germany \\
\texttt{michael.perrot@tuebingen.mpg.de} \\
\And
Ulrike von Luxburg \\
University of T\"{u}bingen, Department of Computer Science, \\
Max-Planck-Institute for Intelligent Systems, \\
T\"{u}bingen, Germany \\
\texttt{ulrike.luxburg@uni-tuebingen.de}
}

\begin{document}

\maketitle

\begin{abstract}
We address the classical problem of hierarchical clustering, but in a framework where one does not have access to a representation of the objects or their pairwise similarities.
Instead, we assume that only a set of comparisons between objects is available, that is, statements of the form ``objects $i$ and $j$ are more similar than objects $k$ and $l$.''
Such a scenario is commonly encountered in crowdsourcing applications.
The focus of this work is to develop comparison-based hierarchical clustering algorithms that do not rely on the principles of ordinal embedding.
We show that single and complete linkage are inherently comparison-based and we develop variants of average linkage.
We provide statistical guarantees for the different methods under a planted hierarchical partition model.
We also empirically demonstrate the performance of the proposed approaches on several datasets.
\end{abstract}

\section{Introduction} \label{introduction}

The definition of clustering as \textit{the task of grouping similar objects} emphasizes the importance of assessing similarity scores for the process of clustering. 
Unfortunately, many applications of data analysis, particularly in crowdsourcing and psychometric problems, do not come with a natural representation of the underlying objects or a well-defined similarity function between pairs of objects.
Instead, one only has access to the results of comparisons of similarities, for instance, quadruplet comparisons of the form ``similarity between $x_i$ and $x_j$ is larger than similarity between $x_k$ and $x_l$.''

The importance and robustness of collecting such ordinal information from human subjects and crowds has been widely discussed in the psychometric and crowdsourcing literature \citep{shepard1962theanalysis,young1987multidimensional,borg2005modern,stewart2005absolute}.
Subsequently, there has been growing interest in the machine learning and statistics communities to perform data analysis in a comparison-based framework 
\citep{agarwal2007generalized,van2012stochastic,heikinheimo2013crowd,zhang2015jointly,arias2017some,haghiri2018comparison}.
The traditional approach for learning in an ordinal setup involves a two step procedure---first obtain a Euclidean embedding of the objects from available similarity comparisons, and subsequently learn from the embedded data using standard machine learning techniques \citep{borg2005modern,agarwal2007generalized,jamieson2011low,tamuz2011adaptively,van2012stochastic,terada2014local,amid2015multiview}.
As a consequence, the statistical performance of the resulting comparison-based learning algorithms relies both on the goodness of the embedding and the subsequent statistical consistency of learning from the embedded data.
While there exists theoretical guarantees on the accuracy of ordinal embedding \citep{jamieson2011low,kleindessner2014uniqueness,jain2016finite,arias2017some}, it is not known if one can design provably consistent learning algorithms using mutually dependent embedded data points.

An alternative approach, which has become popular in recent years, is to directly learn from the ordinal relations. This approach has been used for estimation of data dimension, centroid or density \citep{kleindessner2015dimensionality,heikinheimo2013crowd,ukkonen2015crowdsourced}, 
object retrieval and nearest neighbour search \citep{kazemi2018comparison,haghiri2017comparison},
classification and regression \citep{haghiri2018comparison},
clustering \citep{kleindessner2017kernel,ukkonen2017crowdsourced}, 
as well as hierarchical clustering \citep{vikram2016interactive,emamjomehzadeh2018adaptive}.
The theoretical advantage of a direct learning principle over an indirect embedding-based approach is reflected by the fact that some of the above works come with statistical guarantees for learning from ordinal comparisons \citep{haghiri2017comparison,haghiri2018comparison,kazemi2018comparison}.

\paragraph{Motivation.}
The motivation for the present work arises from the absence of comparison-based clustering algorithms that have strong statistical guarantees, or more generally, the limited theory in the context of comparison-based clustering and hierarchical clustering.
While theoretical foundations of standard hierarchical clustering can be found in the literature \citep{hartigan1981consistency,chaudhuri2014consistent,dasgupta2016cost,moseley2017approximation}, corresponding works in the ordinal setup has been limited \citep{emamjomehzadeh2018adaptive}.
A naive approach to derive guarantees for comparison-based clustering would be to combine the analysis of a classic clustering or hierarchical clustering algorithm with existing guarantees for ordinal embedding \citep{arias2017some}.
Unfortunately, this does not work since the known worst-case error rates for ordinal embedding are too weak to provide any reasonable guarantee for the resulting comparison-based clustering algorithm.
The existing guarantees for ordinal hierarchical clustering hold under a triplet framework, where each comparison returns the two most similar among three objects \citep{emamjomehzadeh2018adaptive}.
The results show that the underlying hierarchy can be recovered by few adaptively chosen comparisons, but if the comparisons are provided beforehand, which is the case in crowdsourcing, then the number of required comparisons is rather large.
The focus of the present work is to develop provable comparison-based hierarchical clustering algorithms that can find an underlying hierarchy in a set of objects given either adaptively or non-adaptively chosen sets of comparisons.

\paragraph{Contribution 1: Agglomerative algorithms for comparison-based clustering.}
The only known hierarchical clustering algorithm in a comparison-based framework employs a divisive approach \citep{emamjomehzadeh2018adaptive}.
We observe that it is easy to perform agglomerative hierarchical clustering using only comparisons since one can directly reformulate single linkage and complete linkage clustering algorithms in the quadruplet comparisons framework.
However, it is well known that single and complete linkage algorithms typically have poor worst-case guarantees \citep{cohenaddad2018hierarchical}.
While average linkage clustering has stronger theoretical guarantees \citep{moseley2017approximation,cohenaddad2018hierarchical}, it cannot be used in the comparison-based setup since it relies on an averaging of similarity scores.
We propose two variants of average linkage clustering that can be applied to the quadruplet comparisons framework.
We numerically compare the merits of these new methods with single and complete linkage and embedding based approaches.

\paragraph{Contribution 2: Guarantees for true hierarchy recovery.}
\citet{dasgupta2016cost} provided a new perspective for hierarchical clustering in terms of optimizing a cost function that depends on the pairwise similarities between objects.
Subsequently, theoretical research has focused on worst-case analysis of different algorithms with respect to this cost function \citep{roy2016hierarchical,moseley2017approximation,cohenaddad2018hierarchical}.
However, such an analysis is complicated in an ordinal setup, where the algorithm is oblivious to the pairwise similarities.
In this case, one can study a stronger notion of guarantee in terms of exact recovery of the true hierarchy \citep{emamjomehzadeh2018adaptive}.
That work, however, considers a simplistic noise model, where the result of each comparison may be randomly flipped independently of other comparisons \citep{jain2016finite}.
Such an independent noise can be easily tackled by repeatedly querying the same comparison and using a majority vote.
It cannot account for noise in the underlying objects and their associated similarities.
Instead, we consider a theoretical model that generates random pairwise similarities with a planted hierarchical structure \citep{balakrishnan2011noise}.
This induces considerable dependence among the quadruplets, and makes the analysis challenging.
We derive conditions under which different comparison-based agglomerative algorithms can exactly recover the hierarchy with high probability.

\section{Background}

In this section we introduce standard hierarchical clustering with known similarities, we describe the model used for the theoretical analyses, and we formalize the comparison-based framework.

\subsection{Agglomerative hierarchical clustering with known similarity scores}\label{subsec:hcstdlinkage}

Let $\setX = \left\lbrace x_i \right\rbrace_{i=1}^N$ be a set of $N$ objects, which may not have a known feature representation.
We assume that there exists an underlying symmetric similarity function $w:\setX\times\setX\to\nsetR$.
The goal of hierarchical clustering is to group the $N$ objects to form a binary tree such that $x_i$ and $x_j$ are merged in the bottom of the tree if their similarity score $w_{ij} = \fw{x_i,x_j}$ is high, and vice-versa.
Here, we briefly review popular agglomerative clustering algorithms \citep{cohenaddad2018hierarchical}.
They rely on the similarity score $w$ between objects to define a similarity function between any two clusters, $W:2^\setX \times 2^\setX \to \nsetR$.
Starting from $N$ singleton clusters, each iteration of the algorithm merges the two most similar clusters.
This is described in Algorithm~\ref{alg:hc}, where different choices of $W$ lead to different algorithms.
Given two clusters $G$ and $G'$, popular choices for $W(G,G')$ are
\begin{align*}
W(G,G') = && && \underbrace{\max_{x_i \in G, x_j \in G'} w_{ij}\text{,}}_\text{\textbf{Single Linkage (SL)}} &&\text{ or }&& \underbrace{\min_{x_i \in G, x_j \in G'} w_{ij}\text{,}}_\text{\textbf{Complete Linkage (CL)}} &&\text{ or }&& \underbrace{\sum_{x_i \in G, x_j \in G'} \frac{w_{ij}}{|G||G'|}\text{.}}_\text{\textbf{Average Linkage (AL)}}
\end{align*}

\begin{algorithm}[t]
\hrule
\SetKwInOut{Input}{input}\SetKwInOut{Output}{output}
\DontPrintSemicolon
\Input{Set of objects $\setX = \left\lbrace x_1,\ldots,x_N \right\rbrace$;
 Cluster-level similarity $W:2^{\setX}\times2^{\setX}\to\nsetR$.}
\Output{Binary tree, or dendrogram, representing a hierarchical clustering of $\setX$.}
\Begin{
Let $\setB$ be a collection of $N$ singleton trees $\setC_1,\ldots, \setC_N$ with root nodes $\setC_i.root = \{x_i\}$.\;
\While{$\left|\setB\right| > 1$}{
Let $\setC,\setC'$ be the pair of trees in $\setB$ for which $W(\setC.root,\setC'.root)$ is maximum.\;
Create $\setC''$ with $\setC''.root = \left\lbrace\setC.root\cup\setC'.root\right\rbrace$, 
$\setC''.left = \setC$, and
$\setC''.right = \setC'$.\;
Add $\setC''$ to the collection $\setB$, and remove $\setC,\setC'$.\;
}
\Return The surviving element in $\setB$.\;
}
\hrule
\caption[AHC]{Agglomerative Hierarchical Clustering.\label{alg:hc}}
\end{algorithm}

\subsection{Planted hierarchical model} \label{subsec:model}

Theoretically, we study the problem of hierarchical clustering under a \emph{noisy hierarchical block matrix} \citep{balakrishnan2011noise} where, given $N$ objects, the matrix containing all the similarities can be written as $M + R$. $M$ is an ideal similarity matrix characterizing the hierarchy among the examples and $R$ is a perturbation matrix that accounts for the noise in the observed similarity scores. In this paper, we assume that the perturbation matrix $R$ is symmetric ($r_{ij} = r_{ji}$ for all $i<j$) and that its entries $\{r_{ij}\}_{1\leq i<j\leq N}$ are mutually independent and normally distributed, that is $r_{ij} \sim \normal{0}{\sigma^2}$, for some fixed variance $\sigma^2$. We define the ideal similarity matrix $M$ as follows. Let $N$ be large and, for simplicity, of the form $N = 2^L N_0$ for integers $L$ and $N_0$. Let $\mu$ be a constant and $\delta>0$. The hierarchical structure, illustrated in Figure~\ref{app:fig:generativemodel} in the appendix, is constructed as follows:
\\\textbf{Step-$0$:} $\setX$ is divided into two equal sized clusters, and, given $x_i$ and $x_j$ lying in different clusters, their ideal similarity is set to $\mu_{ij} = \mu - L\delta$ (dark blue off-diagonal block in Figure~\ref{app:fig:generativemodel}).
\\\textbf{Step-$1$:} Each of the two groups is further divided into two sub-groups, and, for each pair $x_i,x_j$ separated due to this sub-group formation, we set $\mu_{ij} = \mu - (L-1)\delta$.
\\\textbf{Step-$2,\ldots,L-1$:} The above process is repeated $L-1$ times, and in step $\ell$, the ideal similarity across two newly-formed sub-groups is $\mu_{ij} = \mu - (L-\ell)\delta$.
\\\textbf{Step-$L$:} The above steps form $2^L$ clusters, $\setG_1,\ldots,\setG_{2^L}$, each of size $N_0$.
The ideal similarity between two objects $x_i,x_j$ belonging to the same cluster is $\mu_{ij} = \mu$  (yellow blocks in Figure~\ref{app:fig:generativemodel}).

This gives rise to similarities of the form $w_{ij} = \mu_{ij} + r_{ij}$ for all $i<j$. By symmetry of $M$ and $R$, $w_{ji} = w_{ij}$.
We can equivalently assume that, for all $i<j$, the similarities are independently drawn as $w_{ij} = w_{ji} \sim \normal{\mu_{ij}}{\sigma^2}$.
Then, the above process may be viewed as building a hierarchical binary tree, where the clusters $\setG_1,\ldots,\setG_{2^L}$ lie at level-$L$, and are successively merged at higher levels until we reach level-0 (top level) that corresponds to the entire set $\setX$ (see Figure~\ref{app:fig:generativemodel}). 
The pairwise similarity gets smaller in expectation when two objects are merged higher in the true hierarchy.

We consider the problem of exact recovery of the above planted structure, that is correct identification of all the pure clusters $\setG_1,\ldots,\setG_{2^L}$ and recovery of the entire hierarchy among the clusters.

\subsection{The comparison-based framework}
\label{subsec:cbframework}

In Section~\ref{subsec:hcstdlinkage} we assumed that, even without a representation of the objects, we had access to a similarity function $w$.
In the rest of this paper, we consider the ordinal setting, where $w$ is not available, and information about similarities can only be accessed through quadruplet comparisons.
We assume that we are given a set
$
\setQ \subseteq \left\lbrace (i,j,k,l) : x_i,x_j,x_k,x_l \in \setX, w_{ij} > w_{kl} \right\rbrace\text{, }
$
that is, for every ordered tuple $(i,j,k,l) \in \setQ$, we know that $x_i$ and $x_j$ are more similar than $x_k$ and $x_l$.
There exists a total of $\bigO{N^4}$ quadruplets, but in a practical crowdsourcing application, the available set $\setQ$ may only be a small subset of all possible quadruplets.
Since noise is inherent in the similarities, we do not consider it in the comparisons.
We assume $\setQ$ is obtained in either of the two following ways:
\\ \textbf{Active comparisons:} In this case, the algorithm can adaptively ask an oracle quadruplet queries of the form $w_{ij} \gtrless w_{kl}$ and the outcome will be either $w_{ij} > w_{kl}$ or $w_{ij} < w_{kl}$.
\\ \textbf{Passive comparisons:} In this case, for every tuple $(i,j,k,l)$, we assume that with some sampling probability $p\in(0,1]$, there is a comparison $w_{ij} \gtrless w_{kl}$ and based on the outcome either $(i,j,k,l)\in \setQ$ or $(k,l,i,j)\in \setQ$. 
We also assume that the observations of the quadruplets are independent.

\section{Comparison-based hierarchical clustering}
\label{sec:cbhc}

In this section, we discuss that single linkage and complete linkage can be easily implemented in the comparison-based setting, provided that we have access to $\bigOmega{N^2}$ adaptively selected quadruplets.
However, their statistical guarantees are very weak.
It prompts us to study two variants of average linkage.
On the one hand, Quadruplets-based Average Linkage (4--AL) uses average linkage-like ideas to directly estimate the cluster level similarities from the quadruplet comparisons.
On the other hand, Quadruplets Kernel Average Linkage (4K--AL) uses the quadruplet comparisons to estimate the similarities between the different objects and then uses standard average linkage.
We show that both of these variants have good statistical performances in the following senses: (i) they can exactly recover the planted hierarchy under mild assumptions on the signal-to-noise ratio $\frac{\delta}{\sigma}$ and the size of the pure clusters $N_0 = \frac{N}{2^L}$ in the model introduced in Section~\ref{subsec:model}, (ii) 4K--AL only needs $\bigO{N \ln N}$ active comparisons to achieve exact recovery, and (iii) both 4K--AL and 4--AL can achieve exact recovery using only a small subset of passively obtained quadruplets (sampling probability $p \ll 1$).

\subsection{Single linkage (SL) and complete linkage (CL)}

The single and complete linkage algorithms inherently fall in the comparison-based framework.
To see this, first notice that the $\argmax$ and $\argmin$ functions used in these methods only depend on quadruplet comparisons. Although it is not possible to exactly compute the linkage value $W(G,G')$, one can retrieve, in each cluster, the pair of objects that achieve the maximum or minimum similarity.
Then, the knowledge of these optimal object pairs is sufficient since our primary aim is to find the pair of clusters $G,G'$ that maximizes $W(G,G')$ and this can be easily achieved through quadruplet comparisons between the optimal object pairs of every $G,G'$.
This discussion emphasizes that CL and SL fall well in the comparison-based framework when the quadruplets can be adaptively chosen---in order to select pairs with minimum or maximum similarities.
The next proposition, proved in the appendix, bounds the number of active comparisons necessary and sufficient to use SL and CL.
\begin{myprop}[Active query complexity of SL and CL\label{th:activecomplexitysclinkage}]
The SL and CL algorithms require at least $\bigOmega{N^2}$ and at most $\bigO{N^2\ln N}$ number of active quadruplet comparisons.
\end{myprop}

We now state a sufficient condition for exact recovery of the planted model for both SL and CL as well as a matching (up to constant) necessary condition for SL.
The proof is in the appendix.
\begin{myth}[Exact recovery of planted hierarchy by SL and CL\label{th:exactsclinkage}]
Assume that $\eta\in(0,1)$.
If $\frac\delta\sigma \geq 4\sqrt{\ln\left(\frac{N}{\eta}\right)}$, then SL and CL exactly recover the planted hierarchy with probability $1-\eta$.
Conversely, for $\frac\delta\sigma \leq \frac14 \sqrt{\ln \left(\frac{N}{2^L}\right)}$ and large $\frac{N}{2^L}$, SL fails to recover the hierarchy with probability $\frac12$. 
\end{myth}

Theorem~\ref{th:exactsclinkage} implies that a necessary and sufficient condition for exact recovery by single linkage is that the signal-to-noise ratio grows as $\sqrt{\ln N}$ with the number of examples.
This strong requirement raises the question of whether one can achieve exact recovery under weaker assumptions and with less quadruplets.
The subsequent sections provide an affirmative answer to this question.

\subsection{Quadruplets kernel average linkage (4K--AL)}

Average linkage is difficult to cast to the ordinal framework due to the averaging of pairwise similarities, $w_{ij}$, which cannot be computed using only comparisons.
A first way to overcome this issue is to use the quadruplet comparisons to derive some kind of proxies for the similarities $w_{ij}$.
These proxy similarities can then be directly used in the standard formulation of average linkage.
To derive them we use ideas that are close in spirit to the triplet comparisons-based kernel developed by \citet{kleindessner2017kernel}.
Furthermore, we propose two different definitions depending on whether we use active comparisons (Equation~\ref{eq:Kendallkernel}) or passive comparisons (Equation~\ref{eq:Kendallkernel-passive}).

\paragraph{Active case.} We first consider the active case, where the quadruplet comparisons to be evaluated can be chosen by the algorithm.
A pair of distinct items $(i_0,j_0)$ is chosen uniformly at random, and a set of landmark points $\setS$ is constructed such that every $k\in\{1,\ldots,N\}$ is independently added to $\setS$ with probability $q$.
The proxy similarity between two distinct objects $x_i$ and $x_j$ is then defined as
\begin{align}
K_{ij} &= 
\sum_{k\in\setS\backslash\{i,j\}} \left(\mathbb{I}_{\left(w_{ik} > w_{i_0j_0}\right)} - \mathbb{I}_{\left(w_{ik} < w_{i_0j_0}\right)}\right)
\left(\mathbb{I}_{\left(w_{jk} > w_{i_0j_0}\right)} - \mathbb{I}_{\left(w_{jk} < w_{i_0j_0}\right)}\right).
\label{eq:Kendallkernel}
\end{align}
The underlying idea is that two similar objects should behave similarly with respect to any third object, that is if $x_i$ and $x_j$ are similar then we should have $w_{ik} \approx w_{jk}$ for any other object $x_k$.
Since we cannot directly access the similarities, we instead use comparisons to a reference similarity $w_{i_0j_0}$ to evaluate the closeness between $w_{ik}$ and $w_{jk}$.

The next theorem presents exact recovery guarantees for 4K--AL with actively obtained comparisons.
\begin{myth}[Exact recovery of planted hierarchy by 4K--AL with active comparisons\label{th:exact4K--AL}]
Let $\eta\in(0,1)$ and $\Delta = \frac{\eta^2}{100} \frac{\delta}{\sigma} e^{-2L^2\delta^2/\sigma^2}$. 
There exists an absolute constant $C>0$ such that if $N_0 > \frac{4}{\Delta}\sqrt{N}$ and we set
$
    q > \max\left\{C\frac{2^{2L}}{N \Delta^4}\ln \left(\frac{N}{\eta}\right), \frac{3}{N}\ln\left(\frac{2}{\eta}\right)\right\},
$
then with probability at least $1-\eta$, 4K--AL exactly recovers the planted hierarchy using at most $2qN^2$ number of actively chosen quadruplet comparisons.

In particular, if $L=\bigO{1}$, the above statement implies that even with $\frac\delta\sigma$ constant, 4K--AL exactly recovers the planted hierarchy with probability $1-\eta$ using only $\bigO{N\ln N}$ active comparisons.
\end{myth}

The above result shows that, in comparison to SL or CL, the proposed 4K--AL method achieves consistency for smaller signal-to-noise ratio $\frac\delta\sigma$, and can also do so with only $\bigO{N\ln N}$ active comparisons, which is much smaller than that needed by SL and CL. Our result also aligns with the conclusion of~\citet{emamjomehzadeh2018adaptive}, who showed that $\bigO{N\ln N}$ active triplet comparisons suffice to recover hierarchy under a different (data-independent) noise model.

From a theoretical perspective, it is sufficient to use a single random reference similarity $w_{i_0j_0}$. However, in practice, we observe better performances when considering a set $\setR$ of multiple reference pairs. 
Hence, in the experiments, we use the following extension of the above kernel function:
\begin{align}
K_{ij} &= 
\sum_{(i_0,j_0) \in \setR} \sum_{k\in\setS\backslash\{i,j\}} \left(\mathbb{I}_{\left(w_{ik} > w_{i_0j_0}\right)} - \mathbb{I}_{\left(w_{ik} < w_{i_0j_0}\right)}\right)
\left(\mathbb{I}_{\left(w_{jk} > w_{i_0j_0}\right)} - \mathbb{I}_{\left(w_{jk} < w_{i_0j_0}\right)}\right). \label{eq:Kendalkernel-practical}
\end{align}

\paragraph{Passive case.} Theorem~\ref{th:exact4K--AL} shows that 4K--AL can exactly recover the planted hierarchy even for a constant signal-to-noise ratio, provided that it can actively choose the quadruplets.
It is natural to ask if the same holds in the passive case, where we do not have the freedom of querying specific comparisons but instead have access to a small pre-computed set of quadruplet comparisons $\setQ$.
We address this problem using the following variant of the aforementioned quadruplets kernel: 
\begin{align}
K_{ij} &=
\sum_{k,l=1,k<l}^N \sum_{r=1}^{N} \left(\mathbb{I}_{\left(i,r,k,l\right) \in \setQ} - \mathbb{I}_{\left(k,l,i,r\right) \in \setQ}\right) 
\left(\mathbb{I}_{\left(j,r,k,l\right) \in \setQ} - \mathbb{I}_{\left(k,l,j,r\right) \in \setQ}\right)
\label{eq:Kendallkernel-passive}
\end{align}
for all $i\neq j$. This formulation extends the active kernel in~\eqref{eq:Kendallkernel} by using all $\binom{N}{2}$ pairs of $(k,l)$ as reference similarities instead of a single pair $(i_0,j_0)$.
But each term in the sum contributes only when we simultaneously observe the comparisons between $(i,r)$ and $(k,l)$ and between $(j,r)$ and $(k,l)$.
Theorem~\ref{th:exact4K--AL-passive} presents guarantees for 4K--AL with quadruplets obtained from the passive comparisons model in Section~\ref{subsec:cbframework}.
The derived conditions for exact recovery are similar to Theorem~\ref{th:exact4K--AL} in terms of $\frac\delta\sigma$, but passive 4K--AL requires a much larger number of passive comparisons than active 4K--AL.
\begin{myth}[Exact recovery of planted hierarchy by 4K--AL with passive comparisons\label{th:exact4K--AL-passive}]
Let $\eta\in(0,1)$ and $\Delta = \frac{\delta}{2\sigma} e^{-L^2\delta^2/4\sigma^2}$. 
There exists an absolute constant $C>0$ such that if $N_0 > \frac{8}{\Delta}\sqrt{N}$ and we set
$
    p > \max\left\{C\frac{2^{L}}{\Delta^2} \sqrt{\frac1N\ln \left(\frac{N}{\eta}\right)}, \frac{2}{N^4}\ln\left(\frac{2}{\eta}\right)\right\},
$
then with probability at least $1-\eta$, the 4K--AL algorithm exactly recovers the planted hierarchy using at most $pN^4$ quadruplet comparisons, which are passively obtained based on the model described in Section~\ref{subsec:cbframework}.

In particular, if $L=\bigO{1}$, the above statement implies that even with $\frac\delta\sigma$ constant, 4K--AL exactly recovers the planted hierarchy with probability $1-\eta$ using $\bigO{N^{7/2}\ln N}$ passive comparisons.
\end{myth}

\subsection{Quadruplets-based average linkage (4--AL)}

In 4K--AL we derived a proxy for the similarities between objects and then used standard average linkage.
In this section we consider a different approach where we use the quadruplet comparisons to define a new cluster-level similarity function.
This method is particularly well suited when it is not possible to actively query the comparisons.
We assume that we are given a set of passively obtained quadruplets $\setQ$ as in the previous section (4K--AL with passive comparisons).
Using this set of comparisons, one can estimate the relative similarity between two pairs of clusters.
For instance, let $G_1,G_2,G_3,G_4$ be four clusters such that $G_1, G_2$ are disjoint and  so are $G_3,G_4$, and define
\begin{align}
\label{eq:4--AL-pre}
\Wb_\setQ\left(G_1,G_2 \| G_3,G_4\right) = \sum_{x_i\in G_1}\sum_{x_j\in G_2}\sum_{x_k\in G_3} \sum_{x_l\in G_4} \frac{\mathbb{I}_{\left(i,j,k,l\right) \in \setQ} - \mathbb{I}_{\left(k,l,i,j\right) \in \setQ}}{\left|G_1\right|\left|G_2\right|\left|G_3\right|\left|G_4\right|}.
\end{align}
The idea is that clusters $G_1,G_2$ are more similar to each other than $G_3,G_4$ if their objects are, on average, more similar to each other than the objects of $G_3$ and $G_4$. 
This formulation suggests that an agglomerative clustering should merge $G_1,G_2$ before $G_3,G_4$ if $\Wb_\setQ\left(G_1,G_2 \| G_3,G_4\right)>0$.
Also, note that $\Wb_\setQ\left(G_1,G_2 \| G_3,G_4\right) = -\Wb_\setQ\left(G_3,G_4 \| G_1,G_2\right)$ and $\Wb_\setQ\left(G_1,G_2 \| G_1,G_2\right) = 0$,
which hints that~\eqref{eq:4--AL-pre} is a preference relation between pairs of clusters.
We use the above preference relation $\Wb_\setQ$ to define a new cluster-level similarity function $W$ that can be used in Algorithm~\ref{alg:hc}.
Hence, given two clusters $G_p,G_q$, $p\neq q$, we define their similarity as
\begin{align}
\label{eq:4--AL}
W\left(G_p,G_q\right) = \sum_{\substack{r,s=1, r \neq s}}^K \frac{\Wb_\setQ\left(G_p,G_q \| G_r,G_s\right)}{K(K-1)}\,.
\end{align}
The idea is that two clusters $G_p$ and $G_q$ are similar to each other if, on average, the pair is often preferred over the other possible cluster pairs.
The above measure $W$ provides an average linkage approach based on quadruplets (4--AL), whose statistical guarantees are presented below.

\begin{myth}[Exact recovery of planted hierarchy by 4--AL with passive comparisons\label{th:exact4--AL}] 
Let $\eta\in(0,1)$ and $\Delta = \frac{\delta}{2\sigma} e^{-L^2\delta^2/4\sigma^2}$. Assume the following: 
\\
(i) An initial step partitions $\setX$ into pure clusters of sizes in the range $[m,2m]$ for some $m \leq \frac12 N_0$.
\\
(ii) $\setQ$ is a passively obtained set of quadruplet comparisons, where each tuple $(i,j,k,l)$ is observed independently with probability $p > \displaystyle \frac{C}{m\Delta^2} \max\left\{\ln N, \frac1m \ln\left(\frac1\eta\right)\right\}$ for some constant $C>0$.  
\\Then,  with probability $1-\eta$, starting from the given initial partition and using $|\setQ| \leq pN^4$ number of passive comparisons, 4--AL exactly recovers the planted hierarchy.

In particular, if $L=\bigO{1}$, the above statement implies that, when $\frac\delta\sigma$ is a constant, 4--AL exactly recovers the planted hierarchy with probability $1-\eta$ using $\bigO{\frac{N^4\ln N}{m}}$ passive comparisons.
\end{myth}

Compared to 4K--AL (Theorem~\ref{th:exact4K--AL-passive}), the guarantee for 4-AL in Theorem~\ref{th:exact4--AL} additionally requires an initial partitioning of $\setX$ into small pure clusters of size $m$.
This is reasonable in the context of existing hierarchical clustering literature which prove consistency of average linkage under a similar assumption \citep{cohenaddad2018hierarchical}. In principle, one may use passive 4K--AL to obtain these initial clusters.
Theorem~\ref{th:exact4--AL} shows that if the initial clusters are of size $\bigOmega{\ln N}$, then we do not need to observe all quadruplets.
Moreover, if $L=\bigO{1}$ and we have $\bigOmega{N_0}$-sized initial clusters, then the subsequent steps of 4--AL require only $\bigO{N^3\ln N}$ passive comparisons out of the $\bigO{N^4}$ total number of available comparisons. This is less quadruplets than 4K--AL, but it is still large for practical purposes.
It remains an open question whether better sampling rates can be achieved in the passive case.
From a practical perspective, our experiments in Section~\ref{sec:experiments} demonstrate that 4--AL performs better than 4K--AL even when no initial clusters are provided, that is $m=1$. 

\section{Experiments} \label{sec:experiments}

In this section we evaluate our approaches on several problems: we empirically verify our theoretical findings, we compare our methods to ordinal embedding based approaches on standard datasets, and we illustrate their behaviour on a comparison-based dataset.

\subsection{Planted hierarchical model}

\begin{figure}
    \centering
    \subfloat[Proportion $p = 0.01$.\label{fig:p1}]{\includegraphics[height=3.35cm]{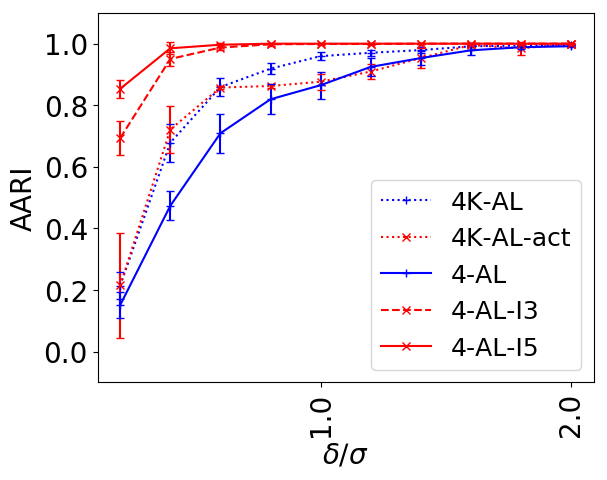}}~~
    \subfloat[Proportion $p = 0.1$.\label{fig:p10}]{\includegraphics[height=3.35cm]{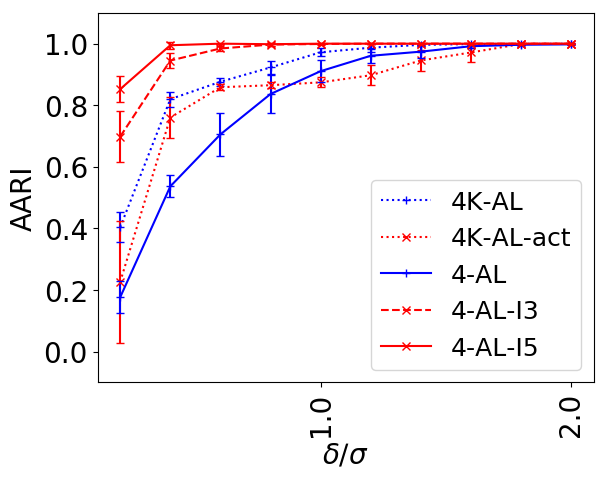}}~~
    \subfloat[Proportion $p = 1$.\label{fig:p100}]{\includegraphics[height=3.35cm]{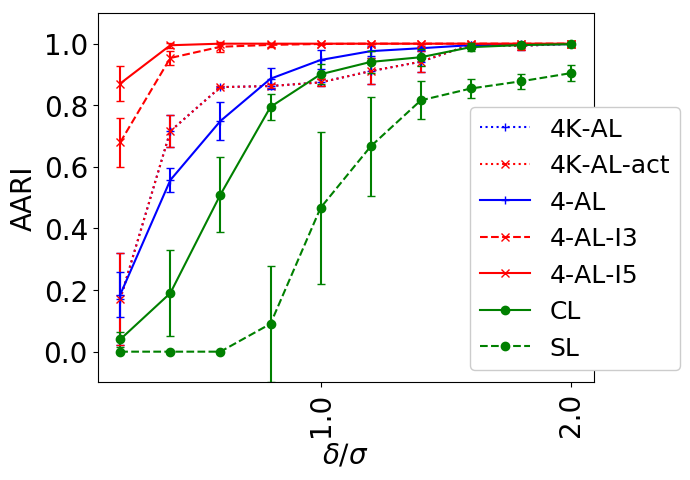}}
	\caption{AARI of the proposed methods (higher is better) on data obtained from the planted hierarchical model with $\mu = 0.8$, $\sigma = 0.1$, $L = 3$, $N_0 = 30$. In Figure~\ref{fig:p1},~\ref{fig:p10},~and,~\ref{fig:p100}, the methods get different proportions $p$ of all the quadruplets. Best viewed in color. 
\label{fig:theoreticalexperimentsAARI}}
\end{figure}

We first use the planted hierarchical model presented in Section~\ref{subsec:model} to generate data and study the performance of the various methods introduced in Section~\ref{sec:cbhc}.

\textbf{Data.}
Recall that our generative model has several parameters, the within-cluster mean similarity $\mu$, the variance $\sigma^2$, the separability constant $\delta$, the depth of the planted partition $L$ and the number of examples in each cluster $N_0$.
From the different guarantees presented in Section~\ref{sec:cbhc}, it is clear that the hardness of the problem depends mainly on the signal-to-noise ratio $\frac{\delta}{\sigma}$, and the probability $p$ of observing samples for the passive methods.
Hence, to study the behaviour of the different methods with respect to these two quantities, we set $\mu = 0.8$, $\sigma = 0.1$, $N_0 = 30$, and $L = 3$ and we vary $\delta \in \left\lbrace 0.02,0.04,\ldots,0.2 \right\rbrace$ and $p \in \left\lbrace 0.01, 0.02, \ldots, 0.1, 1\right\rbrace$.

\textbf{Methods.}
We study SL, CL, which always use the same number of active comparisons and thus are not impacted by $p$.
We also consider 4K--AL with passive comparisons and its active counterpart, 4K--AL--act, implemented as described in~\eqref{eq:Kendalkernel-practical} with $q=\frac{\ln N}{N}$ and the number of references in $\setR$ chosen so that the number of comparisons observed is the same as for the passive methods.
Finally, we study 4--AL with no initial pure clusters and two variants 4--AL--I3 and 4--AL--I5 that have access to initial clusters of sizes 3 and 5 respectively.
These initial clusters were obtained by uniformly sampling without replacement from the $N_0$ examples contained in each of the $2^L$ ground-truth clusters.

\textbf{Evaluation function.}
As a measure of performance we use the Averaged Adjusted Rand Index (AARI) between the ground truth hierarchy and the hierarchies learned by the different methods.
The main idea behind the AARI is to extend the Adjusted Rand Index \citep{hubert1985comparing} to hierarchies by averaging over the different levels (see the appendix for a formal definition).
This measure takes values in $\left[0,1\right]$ with higher values for more similar hierarchies---$\text{AARI}\left(\setC,\setC^\prime\right) = 1$ implies identical hierarchies.
We report the mean and the standard deviation of $10$ repetitions.

\textbf{Results.}
In Figure~\ref{fig:theoreticalexperimentsAARI} we present the results for $p = 0.01$, $p = 0.1$ and $p=1$. We defer the other results to the appendix. Firstly, similar to the theory, SL can hardly recover the planted hierarchy, even for large values of $\frac{\delta}{\sigma}$. CL performs better than SL which implies that it might be possible to derive better guarantees.
We observe that 4K--AL, 4K--AL--act, and, 4--AL are able to exactly recover the true hierarchy for smaller signal-to-noise ratio and their performances do not degrade much when the number of sampled comparisons is reduced.
Finally, as expected, the best method is 4--AL--I5. It uses large initial clusters but recovers the true hierarchy even for very small values of $\frac{\delta}{\sigma}$.

\subsection{Standard clustering datasets}

\begin{figure}
    \centering
    \subfloat[Zoo\label{fig:zoo}]{\includegraphics[height=3.35cm]{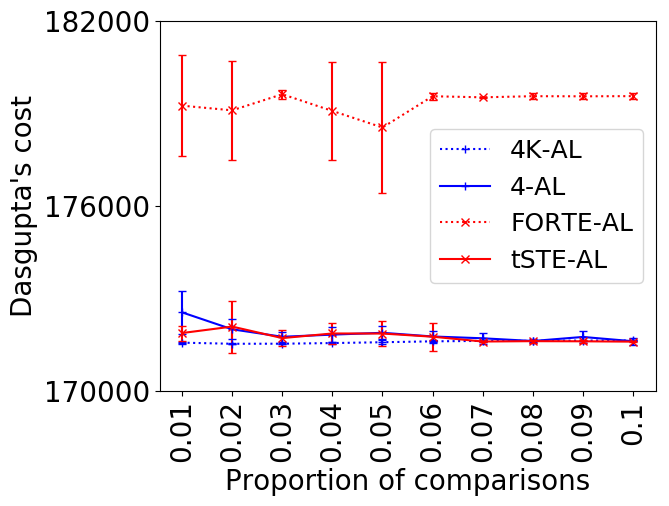}}~~
    \subfloat[Glass\label{fig:Glass}]{\includegraphics[height=3.35cm]{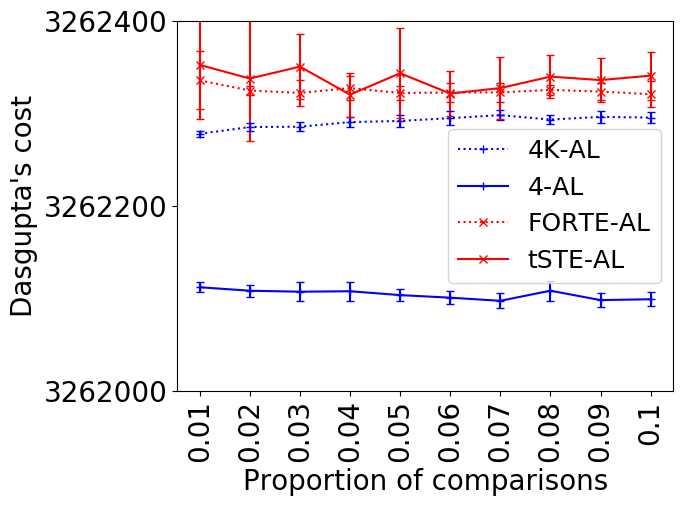}}~~
    \subfloat[20news\label{fig:20news}]{\includegraphics[height=3.35cm]{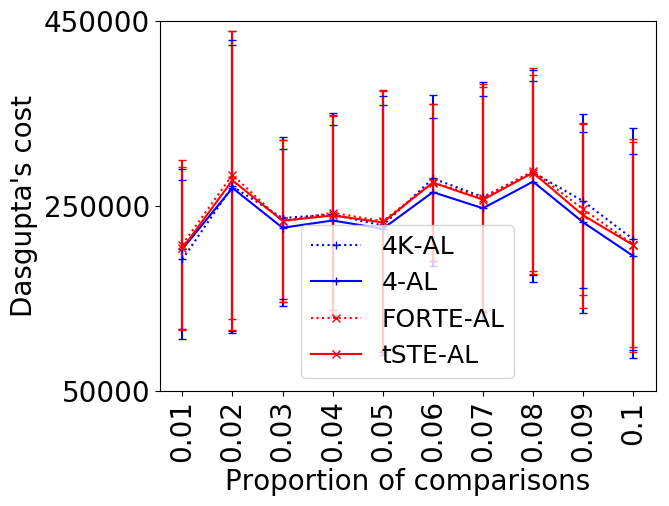}}
	\caption{Dasgupta's score (lower is better) of the different methods on the Zoo, Glass and 20news datasets. The embedding dimension for FORTE--AL and tSTE--AL is set to $2$. Best viewed in color.\label{fig:standarddatasets}}
\end{figure}

In this second set of experiments we compare our passive methods, 4K--AL with passive comparisons and 4--AL without initial clusters, to two baselines that use ordinal embedding as a first step.

\textbf{Baselines.}
We consider t-STE \citep{van2012stochastic} and FORTE \citep{jain2016finite}, followed by a standard average linkage approach using a cosine similarity as the base metric (tSTE-AL and FORTE-AL).
These two methods are parametrized by the embedding dimension $d$.
Since it is often difficult to automatically tune parameters in clustering (because of the lack of ground-truth) we consider several embedding dimensions and report the best results in the main paper.
In the appendix we detail the cosine similarity and report results for other embedding dimensions.

\textbf{Data.}
We evaluate the different approaches on $3$ different datasets commonly used in hierarchical clustering: Zoo, Glass and 20news \citep{heller2005bayesian,vikram2016interactive}.
To fit the comparison-based setting we generate the comparisons using the cosine similarity.
Since it is not realistic to assume that all the comparisons are available. We use the procedure described in Section~\ref{subsec:cbframework} to passively obtain a proportion $p \in \left\lbrace 0.01, 0.02, \ldots, 0.1\right\rbrace$ of all the quadruplets.
Some statistics on the datasets and details on the comparisons generation are presented in the appendix.

\textbf{Evaluation function.}
Contrary to the planted hierarchical model, we do not have access to a ground-truth hierarchy and thus we cannot use the AARI measure.
Instead, we use the recently proposed Dasgupta's cost \citep{dasgupta2016cost} that has been specifically designed to evaluate hierarchical clustering methods.
The idea of this cost is that similar objects that are merged higher in the hierarchy should be penalized.
Hence, a lower cost indicates a better hierarchy. Details are provided in the appendix.
For all the experiments we report the mean and the standard deviation of $10$ repetitions.

\textbf{Results.}
We report the results in Figure~\ref{fig:standarddatasets}.
We note that the proportion of comparisons does not have a large impact as the results are, on average, stable across all regimes.
Our methods are either comparable or better than the embedding-based ones.
They do not need to first embed the examples and thus do not impose a strong Euclidean structure on the data.
The impact of this structure is more or less pronounced depending on the dataset. Furthermore, as illustrated in the appendix, a poor choice of embedding dimension can drastically worsen the results of the embedding methods.

\textbf{Comparison-based dataset.}
In the appendix, we also apply the different methods to a comparison-based dataset called Car \citep{kleindessner2017lens}. 

\section{Conclusion} \label{sec:conclusion}

We investigated the problem of hierarchical clustering in a comparison-based setting.
We showed that the single and complete linkage algorithms (SL and CL) could be used in the setting where comparisons are actively queried, but with poor exact recovery guarantees under a planted hierarchical model.
We also proposed two new approaches based on average linkage (4K--AL and 4--AL) that can be used in the setting of passively obtained comparisons with good guarantees in terms of exact recovery of the planted hierarchy.
An active version of 4K--AL achieves exact recovery using only $\bigO{N\ln N}$ active comparisons.
Empirically, we confirmed our theoretical findings and compared our methods to two ordinal embedding based baselines on standard and comparison-based datasets.

\bibliographystyle{apalike}
\bibliography{bibliography}

\appendix
\renewcommand{\thefigure}{S.\arabic{figure}}
\setcounter{figure}{0}  

\section{The hierarchical model and proof for the theoretical results}
\label{app:sec:proofs}

In this section, we illustrate the planted hierarchical model and we provide detailed proofs of Theorems~\ref{app:th:exactsclinkage}--\ref{app:th:exact4--AL}.

\subsection{Illustration of the planted hierarchical model}

\begin{figure}
    \centering
    \begin{minipage}[c]{.675\linewidth}
    \subfloat{\includegraphics[width=\linewidth]{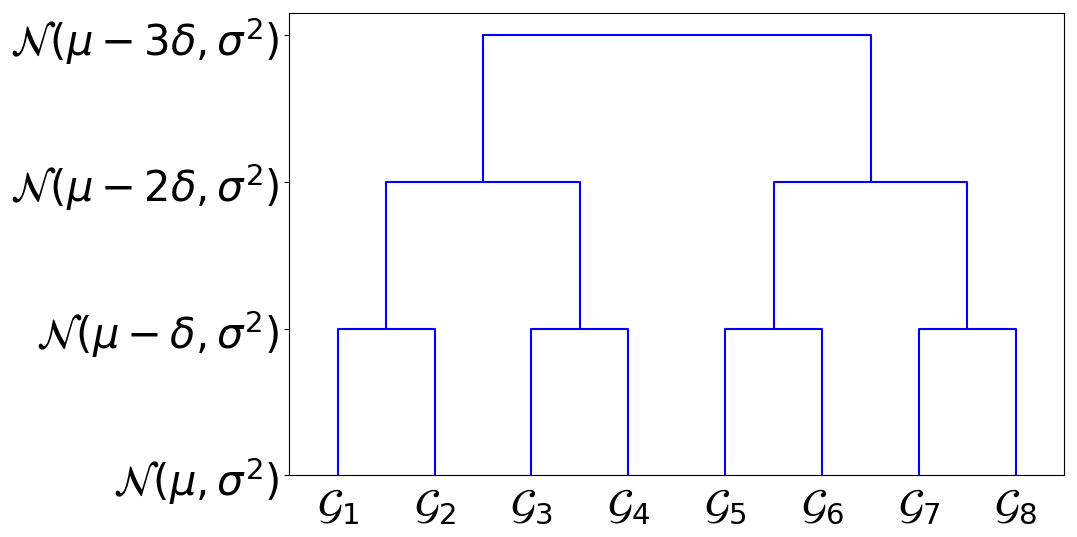}}
    \end{minipage}%
    \hfil
    \begin{minipage}[c]{.325\linewidth}    
    \subfloat{\includegraphics[width=\linewidth]{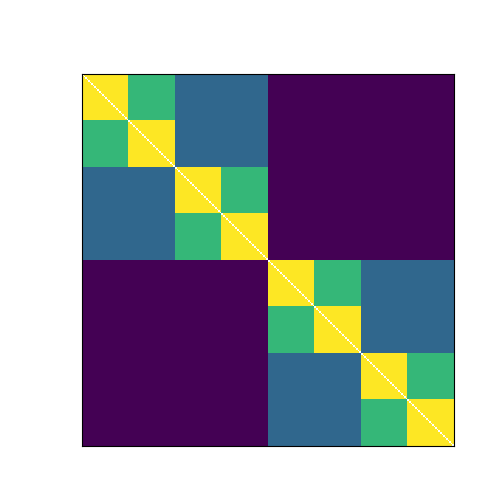}}
    \end{minipage}%
	\caption{\textbf{(Left)} Illustration of the planted hierarchical model for $L=3$ along with specification of the distributions for similarities at different levels; 
	\textbf{(Right)} Hierarchical block structure in the expected pairwise similarity matrix, where darker implies smaller similarity. 
	\label{app:fig:generativemodel}}
\end{figure}

We first illustrate the planted hierarchical model described in Section~\ref{subsec:model} in the main paper.
This is presented in Figure~\ref{app:fig:generativemodel}.
We also recall some of the key quantities associated with the planted model, which include:
\begin{itemize}
    \item $N$, the number of objects;
    \item $L$, the number of levels in the hierarchy;
    \item $N_0 = \frac{N}{2^L}$, the size of the pure clusters;
    \item $\mu$, expected similarity between pairs belonging to a pure cluster;
    \item $\delta$, the separation between the expected similarities across consecutive levels; and
    \item $\sigma$, the standard deviation of the similarities.
\end{itemize}
Throughout the appendix, we use $Z$ to denote a generic standard normal random variable, that is, $Z\sim\normal{0}{1}$.
We also define $\llca_{ij} = \llca(x_i,x_j)$ as the level of the ground truth tree in which the least common ancestor $(lca)$ of $x_i$ and $x_j$ resides. 
We extend this definition to the level of $lca$ of two clusters $G,G'$, denoted by $\llca(G,G')$. If $G,G'$ are both subsets of the same pure cluster, we write $\llca(G,G') = L$. Hence, the range of $\llca$ is $\{0,1,\ldots,L\}$.

\subsection{Analysis of Single Linkage (SL) and Complete Linkage (CL)}

\begin{reprop}{\ref{th:activecomplexitysclinkage}}[Active query complexity of SL and CL\label{app:th:activecomplexitysclinkage}]
The SL and CL algorithms require at least $\bigOmega{N^2}$ and at most $\bigO{N^2\ln N}$ number of active quadruplet comparisons.
\end{reprop}

\begin{proof}
In the first step of SL or CL, the algorithm merges the pair $x_i,x_j$ if $w_{ij} \geq w_{kl}$ for all $k,l \in \{1,\ldots,N\}$. This requires $\binom{N}{2}$ number of ordinal comparisons to find the minimum, and hence, the active query complexity of SL  and CL is at least $\bigOmega{N^2}$.

To prove an upper bound on the active query complexity, it suffices to observe that single/complete linkage only requires a total ordering of the $\binom{N}{2}$ scalar similarities $\{w_{ij} : i<j\}$.
Using a sorting algorithm such as merge-sort, this ordering can be easily obtained from $\bigO{N^2\ln N}$ actively chosen comparisons.
\end{proof}

\begin{reth}{\ref{th:exactsclinkage}}[Exact recovery of planted hierarchy by SL and CL\label{app:th:exactsclinkage}]
Assume that $\eta\in(0,1)$.
If $\frac\delta\sigma \geq 4\sqrt{\ln\left(\frac{N}{\eta}\right)}$, then SL and CL exactly recover the planted hierarchy with probability $1-\eta$.

Conversely, for $\frac\delta\sigma \leq \frac14 \sqrt{\ln \left(\frac{N}{2^L}\right)}$ and large $\frac{N}{2^L}$, SL fails to recover the hierarchy with probability $\frac12$. 
\end{reth}

\begin{proof}
\textbf{We first prove the sufficient condition for exact recovery.}
Let $Z\sim\normal{0}{1}$. It can be easily verified that $\Pb(|Z| \geq t) \leq \sqrt{\frac2\pi}\frac1t \exp(-0.5t^2)$. For $t\geq1$, we may simply bound this by $\exp(-0.5t^2)$.
Now, observe that for every $i\neq j$, $\frac{w_{ij}-\mu_{ij}}{\sigma}\sim\normal01$.
Using this, we can write
\begin{align*}
\Pb\left({\textstyle\bigcup_{i\neq j}} \left\{ | w_{ij} - \mu_{ij} | \geq \frac\delta2\right\} \right) \leq \sum_{i\neq j} \Pb\left( |Z| \geq \frac{\delta}{2\sigma}\right)
\leq N^2 \exp\left(-\frac{\delta^2}{8\sigma^2}\right)
\end{align*}
since $\delta>2\sigma$ under stated condition. The above probability is smaller than $\eta$ for $\delta \geq 4\sigma\sqrt{\ln(\frac{N}{\eta})}$.
Thus, under the stated condition, $|w_{ij}-\mu_{ij}|<\frac\delta2$ for all $i\neq j$.
We now show that the above scenario leads to exact recovery of the hierarchy by single or complete linkage clustering.
Note that
\begin{align*}
\Eb[w_{ij}] = \mu_{ij} = \mu - (L - \llca_{ij}) \delta
\end{align*}
Due to the concentration of the similarity score $w$, we know that $w_{ij}$ lies in the range $\left( \mu - (L - \llca_{ij}) \delta - \frac\delta2, \mu - (L - \llca_{ij}) \delta  + \frac\delta2\right)$ for all $i\neq j$ with probability $1-\eta$. 
Thus, the similarity scores corresponding to the different levels of the ground truth do not overlap, and this ensures that the agglomerative algorithms merge objects or clusters in the same order as prescribed by the ground truth.
For instance, at the first stage, where the goal is to extract the pure clusters, we have 
$w_{ij} > \mu - \frac\delta2$ if $x_i,x_j$ belong to the same pure cluster, and $w_{ij} < \mu - \frac\delta2$ otherwise.
Hence, both single and complete linkage merge objects in the same cluster first before merging objects from different clusters.
The same argument also holds for the subsequent levels and hence, the claim.

\textbf{We now prove the converse statement for SL.} 
We first prove the result for $L=1$. The argument easily extends to $L>1$ from the observation that exact recovery of the entire hierarchy involves exact recovery for pairs of clusters at $L-1$ levels.
For $L=1$, there are two pure clusters, $\setG_1$ and $\setG_2$, that are split at the top level of the true hierarchy.

Recall that single linkage corresponds to a cluster tree on the set of items \citep{chaudhuri2014consistent}.
For any $t\in\mathbb{R}$, we consider the subgraph $G_t$ of the cluster tree with edge set $E_t=\{(i,j): w_{ij}>t\}$.
Observe that $G_t$ is equivalent to a stochastic block model, where
\begin{equation}
\label{app:eq:pf-sc-sbm}
\Pb\big( (i,j)\in E_t\big) = \left\{
    \begin{array}{ll}
        \displaystyle 1-\Phi\left(\frac{t-\mu}{\sigma}\right) & \text{for } i,j \text{ in the same cluster, and}\\ 
        \displaystyle 1-\Phi\left(\frac{t-\mu+\delta}{\sigma}\right) & \text{when } i,j \text{ belong to different clusters}. 
    \end{array}\right.
\end{equation}
Let $p,q$ denote the aforementioned within and inter-cluster edge probabilities in~\eqref{app:eq:pf-sc-sbm}, and recall the bounds on the Gaussian tail
\begin{align}
    \frac{1}{\sqrt{2\pi}} \frac{1}{2x} e^{-x^2/2} 
    < 1 - \Phi(x) <
    \frac{1}{\sqrt{2\pi}} \frac{1}{x} e^{-x^2/2} \,,
    \label{app:eq:gaussian-tail}
\end{align}
which is valid for all $x\geq1$.
Setting $t = \mu + \sigma \sqrt{2\ln N_0}$, it is easy to verify that 
\begin{align*}
    p < \frac{1}{\sqrt{2\pi}}\frac{1}{N_0\sqrt{2\ln N_0}} \qquad\text{and}\qquad 
    q > \frac{1}{\sqrt{2\pi}}\frac{1}{2\left(\sqrt{2\ln N_0} + \frac\delta\sigma\right)} e^{-\left(\sqrt{2\ln N_0} + \frac\delta\sigma\right)^2/2}.
\end{align*}
Assuming $\frac\delta\sigma < \frac14\sqrt{\ln N_0}$, the lower bound on $q$ can be simplified as $q> \frac{1}{10N_0\sqrt{N_0\ln N_0}}$.
Hence, for large enough $N_0$, we have $p<\frac{1}{N_0}$ and $q\gg \frac{\ln N_0}{N_0^2}$. 
Now observe that the two subgraphs of $G_t$ restricted to $\setG_1$ and $\setG_2$, $G_{t|_{\setG_1}}$ and $G_{t|_{\setG_2}}$, are Erd\H{o}s-R{\'e}nyi graphs, each with $N_0$ vertices and edge-probability $p$.
Using a standard result for random graphs \citep[Chapter~8 of][]{blum2018foundations}, we can conclude that both $G_{t|_{\setG_1}}$ and $G_{t|_{\setG_2}}$ are disconnected with high probability for $p<\frac{1}{N_0}$.
Similarly, since $q\gg \frac{\ln N_0}{N_0^2}$, one can conclude that, with high probability, there exist edges between $G_{t|_{\setG_1}}$ and $G_{t|_{\setG_2}}$.
Based on the cluster tree perspective of single linkage~\citep{chaudhuri2014consistent}, the above conclusions about connectivity of $G_{t|_{\setG_1}}$ and $G_{t|_{\setG_2}}$ implies that SL merges items from $\setG_1$ and $\setG_2$ before extracting the pure clusters.
For large enough $N_0$, the probability of this event is greater than $\frac12$.
\end{proof}

\subsection{Analysis of Active Quadruplets Kernel based Average Linkage (4K--AL)}

Recall that the active quadruplet kernel is defined in the following way.
A pair of distinct items $(i_0,j_0)$ is chosen uniformly, and a set of landmark points $\setS$ is constructed such that every $k\in\{1,\ldots,N\}$ is independently added to $\setS$ with probability $q$.
The kernel $K$ is defined as
\begin{align}
K_{ij} &= 
\sum_{k\in\setS\backslash\{i,j\}} \left(\mathbb{I}_{\left(w_{ik} > w_{i_0j_0}\right)} - \mathbb{I}_{\left(w_{ik} < w_{i_0j_0}\right)}\right)
\left(\mathbb{I}_{\left(w_{jk} > w_{i_0j_0}\right)} - \mathbb{I}_{\left(w_{jk} < w_{i_0j_0}\right)}\right)
\label{app:eq:Kendallkernel}
\end{align}
for $i\neq j$. For ease of notation, we introduce the terms $w^* = w_{i_0j_0}$ and $\xi_k = \mathbb{I}_{(k\in\setS)}$.
It follows that $\xi_1,\ldots,\xi_N \sim_{iid} \text{Bernoulli}(q)$ and,
with these notations, we write the kernel function~\eqref{app:eq:Kendallkernel} as
\begin{align*}
    K_{ij} &= \sum_{k\neq i,j} \xi_k\left(2\mathbb{I}_{\left(w_{ik} > w^*\right)} - 1\right)
    \left(2\mathbb{I}_{\left(w_{jk} > w^*\right)} - 1\right),
\end{align*}
where the re-arrangement of indicators are under the planted model assumption since any two similarity scores are distinct with probability 1 due to the Gaussian assumption.

We now restate and prove the exact recovery guarantee for 4K--AL with actively obtained comparisons.

\begin{reth}{\ref{th:exact4K--AL}}[Exact recovery of planted hierarchy by 4K--AL with active comparisons\label{app:th:exact4K--AL}]
Let $\eta\in(0,1)$ and $\Delta = \frac{\eta^2}{100} \frac{\delta}{\sigma} e^{-2L^2\delta^2/\sigma^2}$. 
There exists an absolute constant $C>0$ such that if $N_0 > \frac{4}{\Delta}\sqrt{N}$ and we set
\begin{align*}
    q > \max\left\{C\frac{2^{2L}}{N \Delta^4}\ln \left(\frac{N}{\eta}\right), \frac{3}{N}\ln\left(\frac{2}{\eta}\right)\right\},
\end{align*}
then with probability at least $1-\eta$, 4K--AL exactly recovers the planted hierarchy using at most $2qN^2$ number of actively chosen quadruplet comparisons.

In particular, if $L=\bigO{1}$, the above statement implies that even with $\frac\delta\sigma$ constant, 4K--AL exactly recovers the planted hierarchy with probability $1-\eta$ using only $\bigO{N\ln N}$ active comparisons.
\end{reth}

\begin{proof}
We prove the result by proving the following statements:
\begin{itemize}
    \item the probability that 4K--AL queries more than $2qN^2$ comparisons is at most $\frac\eta2$, and
    \item the probability of not achieving exact recovery is at most $\frac\eta2$.
\end{itemize}

\textbf{To derive the bound on the number of comparisons}, we observe that evaluation of the entire kernel matrix requires quadruplet comparisons of the form $\mathbb{I}_{(w_{ik}>w^*)}$ for all $i=1,\ldots,N$ and $k\in\setS$. 
Hence, the total number of comparisons is $N|\setS|$, which can be bounded by showing that the  size of $\setS$ is at most $2qN$.
This follows from Bernstein’s inequality since
\begin{align*}
    \Pb(|\setS|>2qN) &= \Pb\left(\sum_{k=1}^N \xi_k - qN > qN\right)
    \\&\leq \exp\left(-\frac{q^2N^2}{2Nq(1-q) + \frac23qN}\right)
    \leq \exp\left(-\frac{qN}{3}\right),
\end{align*}
which is bounded by $\frac{\eta}{2}$ since $q > \frac{3}{N}\ln\left(\frac{2}{\eta}\right)$.

\textbf{To derive the exact recovery guarantee}, we analyze the kernel matrix $K$, and also 4K--AL, conditioned on $w^*$.
For this, we need to characterize the behaviour of $w^*$ under the planted model.
Since $w^*$ is the similarity of a randomly chosen pair, one can observe that $w^* \sim \sum\limits_{\ell=0}^L  a_\ell \normal{\mu-\ell\delta}{\sigma^2}$ has a mixture of Gaussian distribution, where the weights $a_0 = \frac{2^L\binom{N_0}{2}}{\binom{N}{2}}$ and $a_\ell = \frac{2^{L+\ell-2}N_0^2}{\binom{N}{2}}$ for $\ell=1,\ldots,L$ are the proportion of similarities corresponding to item pairs merged at level $(L-\ell)$ of the panted hierarchy.
We claim that, with probability $1-\frac\eta4$, 
\begin{equation}
    \mu - L\delta - \sigma\sqrt{2\ln\left(\frac8\eta\right)}
    < w^* < 
    \mu + \sigma\sqrt{2\ln\left(\frac8\eta\right)}\,.
    \label{app:eq:pf-w-ref}
\end{equation}
The bounds follow from the mixture of Gaussian nature of $w^*$ since
\begin{align*}
    \Pb\left(w^* > t \right) 
    &= \sum_{\ell=0}^L a_\ell \Pb\left(\mu - \ell\delta + \sigma Z > t\right)
    \\&\leq \Pb(\mu + \sigma Z > t) \,,
\end{align*}
where we use $Z$ to denote a standard normal random variable. 
Setting $t=\mu + \sigma\sqrt{2\ln\left(\frac8\eta\right)}$ and using the upper bound on Gaussian tail probability \eqref{app:eq:gaussian-tail}, we can bound the above probability by $\frac\eta8$.
A similar argument holds for the lower bound on $w^*$, where the probability of violating the bound is also at most $\frac\eta8$. Hence, the bounds in \eqref{app:eq:pf-w-ref} hold with probability $1-\frac\eta4$.

We next compute the expected kernel matrix~\eqref{app:eq:Kendallkernel} conditioned on the knowledge of $w^*$.
For this, we first define the quantities
\begin{equation}
\begin{aligned}
    \beta_{\ell,w^*} &= 2\Pb_{Z\sim\normal{0}{1}}\left( \mu - (L-\ell)\delta + \sigma Z > w^* \big| w^* \right) - 1, \text{ and}
    \\
    \beta_\ell &= 2\Pb_{Z,Z'\sim\normal{0}{1}}\left( \mu + \sigma Z > \mu - \ell\delta + \sigma Z' \right) - 1
    = 2\Phi \left(\frac{\ell\delta}{\sqrt{2} \sigma}\right) - 1
\end{aligned}
\label{app:eq:pf-beta-def}
\end{equation}
for any $\ell \in \nsetR$ and $w^*\in\nsetR$.
Observe that $\beta_{\ell,w^*} = \Eb\left[ 2\mathbb{I}_{(w_{ij}>w^*)} - 1 \big| w^*\right]$ when $\llca_{ij} = \ell$, whereas $\beta_{\ell} = \Eb\left[ 2\mathbb{I}_{(w_{ij}>w_{kl})} - 1 \right]$ when $\llca_{ij} - \llca_{kl}= \ell$.
In particular, $\beta_0 = 0$.
Based on \eqref{app:eq:pf-beta-def} and the observation that the product terms in~\eqref{app:eq:Kendallkernel} are independent conditioned on $w^*$, we write for any $i\neq j$,
\begin{align*}
    \Eb\left[ K_{ij} \big| w^*\right]
    &= \sum_{k\neq i,j} q \beta_{\llca_{ik},w^*} \beta_{\llca_{jk},w^*}. 
\end{align*}

Recall that, under the planted hierarchy, $\setX$ is partitioned in pure clusters $\setG_1,\ldots, \setG_{2^L}$. We abuse notation to write $\setG_r$ as the set $\{i: x_i\in\setG_r\}$.
In~\eqref{app:eq:Kendallkernel}, observe that each term in the sum depends only on the groups containing $i,j,k$, and hence, we may only compute it for each group and multiply by the number of terms in the group.
If $i,j\in\setG_1$, then $k$ can take only $(N_0-2)$ values in $\mathcal{G}_1$, and $N_0$ values in other groups.
We may perform the entire computation only at group level, and then use a multiplicative factor of $(1\pm\epsilon)$ with $\epsilon = \frac{4}{N_0}$ to account for fluctuations in the number of terms from each group.
Here, $\Eb[K_{ij}|w^*] = (1\pm \epsilon)a$ denotes $(1-\epsilon)a\leq \Eb[K_{ij}|w^*] \leq (1+ \epsilon)a$.
Allowing a fluctuation of $(1\pm\epsilon)$ also helps to ignore the small effect of the case where $(i,k)$ or $(j,k)$ corresponds to $(i_0,j_0)$, that is, the reference pair for which $w^* = w_{i_0j_0}$.
Thus, for $i,j$ such that $i\neq j$ and $\llca_{ij} = \ell$, we have
\begin{align}
 \Eb[K_{ij}| w^*] 
 &= (1\pm \epsilon) q N_0\sum_{r=1}^{2^L}  \beta_{\llca(i,\setG_r),w^*}  \beta_{\llca(j,\setG_r),w^*} 
 \nonumber
 \\&= (1\pm \epsilon) q N_0\sum_{t,t'=0}^{L}  \beta_{t,w^*}\beta_{t',w^*} \#\{r: \llca(i,\setG_r)=t,\llca(j,\setG_r)=t'\},
  \label{app:eq:pf-EKw-compute}
\end{align}
where the second equality explicitly mentions that we need to count the number of different pure clusters that are merged with $i$ or $j$ at different levels of the true hierarchy.
We now consider different cases. First, if $i,j$ belong to same group, then $\ell=L$ and  $\llca(i,\setG_r)=\llca(j,\setG_r)$  for every $r$. So,
\begin{align*}
    \kappa_L := \Eb[K_{ij}| w^*] 
    &= (1\pm \epsilon) q N_0\sum_{t=0}^{L}  (2^{L-1-t}\vee 1) \beta_{t,w^*}^2,
\end{align*}
which we denote by a quantity $\kappa_L$ noting that it only depends on the level $L$ and not on $i,j$. 
Here, $\vee$ denotes the maximum of two values.
The numbers of clusters are computed based on the fact that there is only one cluster at levels $L$ or $L-1$, and otherwise $2^{L-1-t}$ groups are merged with $i$ at level-$t$.
If $i,j$ are not in the same group, that is, $\ell = \llca_{ij} < L$, then we observe:
\begin{itemize}
    \item if $t<\ell$, then for any $\setG_r$ such that $\llca(i,\setG_r)=t$, we also have $\llca(j,\setG_r)=t$. So we may only consider cases $t=t'$ when $t<\ell$.
    \item there is no $\setG_r$ such that $\llca(i,\setG_r)=\llca(j,\setG_r)=\ell$ which happens because the hierarchy is a binary tree and $\setG_r$ must either merge first with $i$ or with $j$.
    So, we do not need to consider $t=t'=\ell$, which is the main difference from the case $\ell=L$.
    \item if $t>\ell$, then for any $\setG_r$ with $\llca(i,\setG_r)=t$, we have $\llca(j,\setG_r)=\ell$. So we may set $t'=\ell$ whenever we have $t>\ell$.
    Similarly, we should also count the cases $t'>\ell,t=\ell$.
\end{itemize}
Thus, we can decompose the summation into three parts based on the conditions on $t,t'$ ($t=t'<\ell$; $t>\ell, t'=\ell$; $t=\ell,t'>\ell$). 
For each case, we should count $\#\{r:\llca(i,\setG_r)=t,\llca(j,\setG_r)=t'\}$.
To compute these, we note that $\#\{r:\llca(i,\setG_r)=\llca(j,\setG_r)=t\} = 2^{L-1-t}$ when $t=t'<\ell$ as used above.
But when $t>\ell,t'=\ell$, we have $\#\{r:\llca(i,\setG_r)=t,\llca(j,\setG_r)=\ell\} = 2^{L-1-t}\vee1$ since this counts only those groups which merge with $i$ at level-$t$, and $t'$ plays no role in the count.
A similar argument holds for the case $t=\ell,t'>\ell$.
Based on this, we compute $\Eb[K_{ij}|w^*]$ for the case  $\llca_{ij} = \ell <L$ and denote the expected value by $\kappa_\ell$, noting that it does not depend on $i,j$. We have
\begin{align*}
    \kappa_\ell := \Eb[K_{ij} | w^*]
    &= (1\pm\epsilon)q N_0 \left[ 
    \sum_{t=0}^{\ell-1} 2^{L-1-t}\beta_{t,w^*}^2 + 2 \sum_{t=\ell+1}^L (2^{L-1-t}\vee 1)\beta_{t,w^*}\beta_{\ell,w^*} \right],
\end{align*}
where the second term, counted twice, corresponds to both the cases of $t>\ell$ or $t'>\ell$, which behave similarly.
Since $\beta_{t,w^*} \in[-1,1]$, one can easily verify that $|\kappa_\ell| \leq qN$ for all $\ell$.

The above discussion leads to the conclusion that $\Eb[K|w^*]$ has a block diagonal structure with exactly $L+1$ distinct off-diagonal entries, $\kappa_0,\ldots,\kappa_L$, and the block structure corresponds to the planted hierarchy shown in Figure~\ref{app:fig:generativemodel} (right). 
We now show that these distinct terms are sufficiently separated, that is, $\kappa_{\ell+1}-\kappa_\ell$ is large for every $\ell = 0,1,\ldots,L-1$. To derive this, we require a lower bound on  
\begin{align*}
    \beta_{t+1,w^*}-\beta_{t,w^*}
    &= 2\Pb\left( \mu - (L-t-1)\delta + \sigma Z > w^* \big| w^* \right) - 2\Pb\left( \mu - (L-t)\delta + \sigma Z > w^* \big| w^* \right)
    \\&= \sqrt{\frac2\pi} \int\limits_{(w^*-\mu +(L-t-1)\delta)/\sigma}^{(w^*-\mu +(L-t)\delta)/\sigma} e^{-z^2/2}\mathrm{d}z
    \\&\geq \sqrt{\frac2\pi} \frac\delta\sigma e^{-\left( a^2 \vee (a-\delta)^2 \right) /2\sigma^2},
\end{align*}
where $a = w^* - \mu +(L-t)\delta$.
Conditioned on the bounds $w^*$ stated in~\eqref{app:eq:pf-w-ref},
one can see that 
\begin{align*}
    a^2 \vee (a-\delta)^2 < 2(L+1)^2\delta^2 + 4\sigma^2\ln\left(\frac8\eta\right),
\end{align*}
where we use the fact that $t\in[0,L]$ and the inequality $(x+y)^2 \leq 2(x^2+y^2)$. Plugging this into the above derivation shows that $\beta_{t+1,w^*}-\beta_{t,w^*} > \Delta$ for any $t\in[0,L]$, where $\Delta$ is defined in the statement of theorem.
We use the above bound to show that
\begin{align*}
    \kappa_L - \kappa_{L-1}
    &> qN_0 \left(\beta_{L,w^*}^2 - \beta_{L-1,w^*}^2\right)^2 - 2\epsilon qN
    \\&> qN_0 \Delta^2 - q 2^{L+3},
\end{align*}
where the second term, involving $\epsilon$, takes care of the
fluctuation due to our approximate computations of $\kappa_\ell$ and is simply bounded by the upper bound on $\kappa_\ell$.
Similarly, for any $\ell < L-1$,
\begin{align*}
    \kappa_{\ell+1} - \kappa_\ell
    &> qN_0 \bigg[ 2^{L-1-\ell}\beta_{\ell,w^*}^2 - 2^{L-1-\ell}\beta_{\ell,w^*}\beta_{\ell+1,w^*} 
    \\& \hskip25ex + 2\sum_{t=\ell+2}^L (2^{L-1-t}\vee 1) \beta_{t,w^*}(\beta_{\ell+1,w^*} - \beta_{\ell,w^*})\bigg] - 2\epsilon qN
    \\&= qN_0 2\sum_{t=\ell+2}^L (2^{L-1-t}\vee 1) (\beta_{t,w^*}-\beta_{\ell,w^*})(\beta_{\ell+1,w^*} - \beta_{\ell,w^*}) - 2\epsilon qN
    \\&> 2^{L-\ell-1}qN_0 \Delta^2 - q2^{L+3},
\end{align*}
where the equality follows since $2^{L-1-\ell}=2\sum\limits_{t=\ell+2}^{L} (2^{L-1-t}\vee 1)$, and subsequently, we note that $\beta_{t,w^*}-\beta_{\ell,w^*} > \beta_{\ell+1,w^*}-\beta_{\ell,w^*}>\Delta$ for all $t\geq \ell+2$.
Hence, we can conclude that for $N_0 > \frac{4}{\Delta}\sqrt{N}$, or equivalently, $N_0 > \frac{2^{L+4}}{\Delta^2}$, 
\begin{align}
    \kappa_{\ell+1} - \kappa_\ell > \frac{qN_0 \Delta^2}{2}
    \label{app:eq:pf-4kactive-kappa-diff}
\end{align}
for all $\ell = 0,1,\ldots,L-1$.
We subsequently show that under the condition on $q$ assumed in the theorem, with probability $1-\frac\eta4$,
\begin{align}
    K_{ij} - \Eb[K_{ij}|w^*] < \frac{qN_0 \Delta^2}{4}
    \label{app:eq:pf-4kactive-conc}
\end{align}
for all $i\neq j$. 
This implies that all random entries of $K$ corresponding to different levels of hierarchy in the ground truth tree are non-overlapping. 
Hence, one can simply use the arguments in the proof of Theorem~\ref{app:th:exactsclinkage} to show that average linkage (or even single/complete linkage) recovers the planted hierarchy. 
We complete the proof by deriving the concentration result of~\eqref{app:eq:pf-4kactive-conc}.
From~\eqref{app:eq:Kendallkernel}, we observe that, conditioned on $w^*$, the entry $K_{ij}$ is a sum of $N-2$ independent random variables each lying in the range $[-1,1]$.
Hence, a direct application of Bernstein's inequality implies that
\begin{align*}
    \Pb\left( \big| K_{ij} - \Eb[K_{ij}|w^*]\big| > \sqrt{3qN\ln\left(\frac{4N^2}{\eta}\right)} \bigvee 3\ln\left(\frac{4N^2}{\eta}\right) \bigg| w^* \right) \leq \frac{\eta}{2N^2} \,.
\end{align*}
Using the symmetry of $K$ and the union bound, it follows that the above entry-wise concentration holds for all $i\neq j$ with probability at least $1-\frac\eta4$.
Finally, for $q > C\frac{2^{2L}}{N \Delta^4}\ln \left(\frac{N}{\eta}\right)$ with $C>0$ large enough, it is easy to verify that $\frac14 qN_0\Delta^2$ is larger than the deviation obtained using Bernstein's inequality.
The above argument leads to the claim of Theorem~\ref{app:th:exact4K--AL}. 

\textbf{To verify the claim for fixed $L$ and $\frac\delta\sigma$}, we note that, in this case, $\Delta$ is constant and $N_0 = \bigOmega{N}$.
Hence, using $q= \frac{c\ln N}{N}$ for a large enough constant $c$ immediately leads to the exact recovery guarantee and number of comparisons.
\end{proof}

\subsection{Analysis of Passive Quadruplets Kernel based Average Linkage (4K--AL)}

In the passive setting, we do not have the freedom of querying specific comparisons but have access to only a pre-computed set of quadruplet comparisons $\setQ \subset\{(i,j,k,l): w_{ij}> w_{kl}\}$.
Hence, we use a variant of the kernel in~\eqref{app:eq:Kendallkernel}, which relies only on passively obtained comparisons. 
\begin{align}
K_{ij} &= 
\sum_{\substack{k,l=1 \\ k<l}}^N \sum_{r=1}^{N} \left(\mathbb{I}_{\left(i,r,k,l\right) \in \setQ} - \mathbb{I}_{\left(k,l,i,r\right) \in \setQ}\right) 
\left(\mathbb{I}_{\left(j,r,k,l\right) \in \setQ} - \mathbb{I}_{\left(k,l,j,r\right) \in \setQ}\right).
\label{app:eq:Kendallkernel-passive}
\end{align}
In principle, the above kernel extends the actively computed kernel~\eqref{app:eq:Kendallkernel} by using all $\binom{N}{2}$ pairs of $(k,l)$ as references in comparison to only one used in~\eqref{app:eq:Kendallkernel}.
However, each term in the sum only contributes when we simultaneously observe the comparisons between $(i,r)$ and $(k,l)$ and between $(j,r)$ and $(k,l)$.

In the following, we assume that the model for obtaining passive comparisons is the one described in Section~\ref{subsec:cbframework} of the main paper.
For every tuple $(i,r,k,l)$, we assume that with probability $p\in(0,1]$, there is a comparison $w_{ir} \gtrless w_{kl}$ and based on the comparison either $(i,r,k,l)\in \setQ$ or $(k,l,i,r)\in \setQ$. 
We also assume that the observation of the quadruplet comparisons are independent.
Based on this model, we define a set of i.i.d. Bernoullis $\{\xi_{irkl} \sim \text{Bernoulli}(p) : i,r,k,l \text{ such that } i< r, k<l, (i,r)<(k,l)\}$, where we order the indices/ index pairs to avoid repeated counting of the same tuple.
It follows that $|\setQ| = \sum\limits_{i,r,k,l} \xi_{irkl}$, and from Bernstein's inequality, it follows that $|\setQ| = \bigO{pN^4}$ with high probability.
Using this notation, we may re-write the kernel function in~\eqref{app:eq:Kendallkernel-passive} as
\begin{align*}
K_{ij} &= 
\sum_{k<l} \sum_{r\neq i,j} \xi_{irkl}\xi_{jrkl} 
\left(\mathbb{I}_{(w_{ir}>w_{kl})} - \mathbb{I}_{(w_{ir}<w_{kl})}\right) 
\left(\mathbb{I}_{(w_{jr}>w_{kl})} - \mathbb{I}_{(w_{jr}<w_{kl})}\right).
\end{align*}

We now restate and prove the exact recovery guarantee for average linkage with the aforementioned kernel.
\begin{reth}{\ref{th:exact4K--AL-passive}}[Exact recovery of planted hierarchy by 4K--AL with passive comparisons\label{app:th:exact4K--AL-passive}]
Let $\eta\in(0,1)$ and $\Delta = \frac{\delta}{2\sigma} e^{-L^2\delta^2/4\sigma^2}$. 
There exists an absolute constant $C>0$ such that if $N_0 > \frac{8}{\Delta}\sqrt{N}$ and we set
\begin{align*}
    p > \max\left\{C\frac{2^{L}}{\Delta^2} \sqrt{\frac1N\ln \left(\frac{N}{\eta}\right)}, \frac{2}{N^4}\ln\left(\frac{2}{\eta}\right)\right\},
\end{align*}
then with probability at least $1-\eta$, the 4K--AL algorithm exactly recovers the planted hierarchy using at most $pN^4$ quadruplet comparisons, which are passively obtained based on the model described in Section~\ref{subsec:cbframework} (of the main paper).

In particular, if $L=\bigO{1}$, the above statement implies that even with $\frac\delta\sigma$ constant, 4K--AL exactly recovers the planted hierarchy with probability $1-\eta$ using $\bigO{N^{7/2}\ln N}$ passive comparisons.
\end{reth}

\begin{proof}
\textbf{The upper bound on the number of comparisons} follow by noting that $|\setQ|$ is a sum of $\binom{\binom{N}{2}}{2}$ i.i.d. Bernoullis, and hence, the bound of $pN^4$ holds with probability $1-\frac\eta2$ for $p> \frac{2}{N^4} \ln(\frac2\eta)$.

\textbf{The proof for exact recovery} has a similar structure as that of Theorem~\ref{app:th:exact4K--AL}, the only difference being that the analysis does not depend on a fixed reference pair.
In particular, we can write the expected entries of the kernel matrix in~\eqref{app:eq:Kendallkernel-passive} as
\begin{align*}
    \Eb[K_{ij}] &= \sum_{k<l} \sum_{r\neq i,j} p^2 \big(2\Pb(w_{ir}>w_{kl}) - 1\big) \big(2\Pb(w_{jr}>w_{kl}) - 1\big) 
    \\&= \frac12 \sum_{k\neq l} \sum_{r\neq i,j} p^2 \beta_{\llca_{ir}-\llca_{kl}} \beta_{\llca_{jr}-\llca_{kl}},
\end{align*}
where $\beta$ is defined in~\eqref{app:eq:pf-beta-def}.
As in the proof of Theorem~\ref{app:th:exact4K--AL}, we show that $\Eb[K_{ij}]$ can take at most $L+1$ distinct values depending on the level $\llca_{ij}$.
As before, we decompose the above summation depending on $\llca_{ir},\llca_{jr}$ and $\llca_{kl}$, and also allow a fluctuation of $(1\pm \epsilon)$ with $\epsilon = \frac{8}{N_0}$ to take care of minor effects of ignoring cases such as $k=l$ or $r=i,j$.
We  write the expectation in terms of the clusters as
\begin{align*}
    \Eb[K_{ij}] 
    &= \frac{(1\pm \epsilon)}{2} p^2 N_0^3 \sum_{r,k,l=1}^{2^L}  \beta_{\llca(i,\setG_r)-\llca(\setG_k,\setG_l)} \beta_{\llca(j,\setG_r)-\llca(\setG_k,\setG_l)}
    \\&= \frac{(1\pm \epsilon)}{2} p^2 N_0^3 \sum_{s,t,t'=0}^L  \beta_{t-s} \beta_{t'-s} \times 
    \\& \qquad\qquad\qquad\qquad \#\{r: \llca(i,\setG_r) = t, \llca(j,\setG_r) = t'\}
    \#\{k,l: \llca(\setG_k,\setG_l) = s\}
    \\&= (1\pm \epsilon) p^2 N_0^3 2^{L-1} \sum_{s,t,t'=0}^L (2^{L-1-s}\vee 1)  \beta_{t-s} \beta_{t'-s} \#\{r: \llca(i,\setG_r) = t, \llca(j,\setG_r) = t'\}
\end{align*}
The last step holds since every cluster $\setG_l$ is merged with $(2^{L-1-s}\vee1)$ clusters at level-$s$, and hence, $\#\{k,l: \llca(\setG_k,\setG_l) = s\} = 2^L(2^{L-1-s}\vee 1)$.

We now compute $\kappa_\ell = \Eb[K_{ij}]$ where $\ell = \llca_{ij}$.
For, $\ell=L$, that is, when $i,j$ belong to the same cluster, $\llca(i,\setG_r) =  \llca(j,\setG_r)$ for every cluster. Hence,
\begin{align*}
    \kappa_L = (1\pm \epsilon) p^2 N_0^3 2^{L-1} \sum_{s,t=0}^L (2^{L-1-s}\vee 1) (2^{L-1-t}\vee 1)  \beta_{t-s}^2 \,.
\end{align*}
For $\llca_{ij} = \ell < L$, we have three possible cases as mentioned in the proof of Theorem~\ref{app:th:exact4K--AL}: $(t=t'<\ell)$; $(t>\ell, t'=\ell)$; and $(t=\ell, t'>\ell)$.
Decomposing the summation based on these cases and noting that  $(t>\ell, t'=\ell)$ and $(t=\ell, t'>\ell)$ lead to similar terms, we have
\begin{align*}
    \kappa_{\ell} = (1\pm \epsilon) p^2 N_0^3 2^{L-1} \sum_{s=0}^L (2^{L-1-s}\vee 1) \left[ \sum_{t=0}^{\ell-1} 2^{L-1-t} \beta_{t-s}^2 + 2\sum_{t=\ell+1}^L (2^{L-1-t}\vee 1) \beta_{t-s}  \beta_{\ell-s} \right]
\end{align*}
for every $\ell = 0,1,\ldots,L-1$.
We now derive a lower bound on the separation $\kappa_{\ell+1} - \kappa_\ell$,
which depends on the observation that $|\kappa_\ell| \leq \frac12 p^2N^3$ for every $\ell$, and a lower bound on
\begin{align*}
    \beta_{t+1-s} - \beta_{t-s} 
    &\geq \min_{r\in[-L,L-1]} \beta_{r+1} - \beta_r
    \\&= \min_{r\in[-L,L-1]} \sqrt{\frac2\pi} \int\limits_{r\delta/\sqrt{2}\sigma}^{(r+1)\delta/\sqrt{2}\sigma} e^{-z^2/2}\mathrm{d}z
    \\&> \frac{1}{\sqrt{\pi}} \frac\delta\sigma e^{-L^2\delta^2/4\sigma^2} .
\end{align*}
The lower bound is larger than $\Delta$ stated in the theorem.
Based on this bound and noting that $2^L = \sum\limits_{s=0}^L (2^{L-1-s}\vee 1)$, we obtain 
\begin{align*}
    \kappa_L - \kappa_{L-1} 
    &> p^2 N_0^3 2^{L-1} \sum_{s=0}^L (2^{L-s-1}\vee1) (\beta_{L-s} - \beta_{L-1-s})^2 - \epsilon p^2N^3
    \\&> \frac{1}{2^{L+1}} p^2 N^3\Delta^2 - p^2 N^2 2^{L+3},
\end{align*}
which is at least $\frac{1}{2^{L+2}} p^2 N^3 \Delta^2$ if $N > \frac{2^{2L+5}}{\Delta^2}$, or equivalently, $N_0 > \frac{4\sqrt{2}}{\Delta}\sqrt{N}$.
Similarly, for $\ell<L-1$, we have
\begin{align*}
    \kappa_{\ell+1} - \kappa_\ell 
    &> p^2 N_0^3 2^{L-1} \sum_{s=0}^L (2^{L-1-s}\vee1) \bigg[ 2^{L-1-\ell} \beta_{\ell-s}^2 - 2^{L-1-\ell}\beta_{\ell+1-s}\beta_{\ell-s} 
    \\&\hskip25ex + 2\sum_{t=\ell+2}^L (2^{L-1-t}\vee1) \beta_{t-s} (\beta_{\ell+1-s} - \beta_{\ell-s})\bigg] - \epsilon p^2N^3
    \\&> p^2 N_0^3 2^L \sum_{s=0}^L \sum_{t=\ell+2}^L (2^{L-1-s}\vee1) (2^{L-1-t}\vee1) \Delta^2 -  p^2N^2 2^{L+3}
    \\&= p^2 N_0^3 2^{3L-\ell-2} \Delta^2 - p^2N^2 2^{L+3}
    \\&> \frac{p^2 N^3 \Delta^2}{2^{\ell+2}} \,.
\end{align*}
The second step follows by using $2\sum\limits_{t=\ell+2}^L (2^{L-1-t}\vee1) = 2^{L-1-\ell}$ and $\beta_{\ell+1-s} - \beta_{\ell-s} >\Delta$.
The third step computes the summation, and the fourth holds when $N_0 > \frac{8}{\Delta}\sqrt{N}$.
Thus for every $\ell$, we obtain a minimum separation 
\begin{align*}
    \kappa_{\ell+1} - \kappa_\ell > \frac{1}{2^{L+2}} p^2 N^3 \Delta^2.
\end{align*}

Following the proof idea of Theorem~\ref{app:th:exact4K--AL}, it only remains to show that the fluctuation of $|K_{ij}-\Eb[K_{ij}]|$ is less than half of this minimum separation for all $i<j$ since, under this scenario, one can argue that entries of $K$ corresponding to different levels of the planted hierarchy are well-separated, and hence, the planted hierarchy is exactly recovered by average linkage.
Thus, to complete the proof, we derive the following concentration inequality
\begin{align}
    \Pb\left( \big|K_{ij} - \Eb[K_{ij}] \big| > \sqrt{2p^2 N^5\ln\left(\frac{2N^4}{\eta}\right)} \bigvee 2N^2 \ln\left(\frac{2N^4}{\eta}\right) \right) \leq \frac{\eta}{N^2} \,.
    \label{app:eq:pf-4kpassive-conc}
\end{align}
By union bound, it follows that with probability $1-\frac\eta2$, the above bound holds for all $i<j$, whereas setting $p > C\frac{2^{L}}{\Delta^2} \sqrt{\frac1N\ln \left(\frac{N}{\eta}\right)}$ for $C>0$ large enough ensures that the deviation is smaller than $\frac{1}{2^{L+3}} p^2 N^3 \Delta^2$.
To derive~\eqref{app:eq:pf-4kpassive-conc}, we note that $K_{ij} - \Eb[K_{ij}] = \sum\limits_{k<l}\sum\limits_{r\neq i,j} B_{rkl}$ is a sum of $\binom{N}{2}(N-2)$ random variables, where we use $B_{rkl}\in[-1,1]$ to denote each term in the summation.
One can verify that $B_{rkl}$ has zero mean and its variance is smaller that $p^2$.
Moreover, each $B_{rkl}$ is dependant on all $(N-3)$ random variables, $\{B_{r'kl}: r'\neq r\}$, and all $\binom{N}{2}-1$ random variables, $\{B_{rk'l'}:(k',l')\neq (k,l)\}$.
Hence, if we draw a dependency graph among these random variable, we obtain a regular graph with the vertex degree of each node being $(N + \binom{N}{2} - 4) < N^2$.
We use the concentration technique described in Section~2.3.2 of \citet{janson2002infamous}, where the key observation is that for any graph with maximum degree $d$, one can find an equitable colouring with $d+1$ colours, that is a colouring where all colour classes (independent sets) differ in size by at most one.
In the present context, it implies that one can split the set of random variables into at most $N^2$ subsets, $\setC_1,\ldots,\setC_{N^2}$ such that each subset contains at most $\frac{\binom{N}{2}(N-3)}{N^2} < \frac{N}{2}$ variables that are mutually independent.
Hence, we can apply union bound followed by Bernstein's inequality to write
\begin{align*}
    \Pb\left( \big|K_{ij} - \Eb[K_{ij}] \big| > \tau\right) &\leq \Pb\left( \bigcup_{s=1}^{N^2}\left| \sum_{(r,k,l)\in \setC_s} B_{rkl} \right| > \frac{\tau}{N^2}\right)
    \\&\leq \sum_{s=1}^{N^2} \Pb\left( \left| \sum_{(r,k,l)\in \setC_s} B_{rkl} \right| > \frac{\tau}{N^2}\right)
    \\&\leq 2N^2 \exp\left(- \frac{\frac{\tau^2}{N^4}}{p^2N + \frac23\frac{\tau}{N^2}}\right)
    \leq 2N^2 \exp\left(- \frac{\tau^2}{2p^2N^5}\bigvee \frac{\tau}{2N^2}\right).
\end{align*}
For $\tau = \sqrt{2p^2 N^5\ln\left(\frac{2N^4}{\eta}\right)} \bigvee 2N^2 \ln\left(\frac{2N^4}{\eta}\right)$, the probability is smaller than $\frac{\eta}{N^2}$, which results in the conclusion of~\eqref{app:eq:pf-4kpassive-conc}.

\textbf{To verify the claim for fixed $L$ and $\frac\delta\sigma$}, we note that in this case, $\Delta$ is constant and $N_0 = \bigOmega{N}$.
Hence, using $p= c\sqrt{\frac{\ln N}{N}}$ for a large enough constant $c$ immediately leads to the exact recovery guarantee and the number of passive comparisons.
\end{proof}

\subsection{Analysis of Quadruplets based Average Linkage (4--AL)}

The proposed 4--AL algorithms estimates the relative similarity between two pairs of clusters.
For instance, let $G_1,G_2,G_3,G_4$ be four clusters such that $G_1, G_2$ are disjoint and  so are $G_3,G_4$, we define
\begin{align}
\label{app:eq:4--AL-pre}
\Wb_\setQ\left(G_1,G_2 \| G_3,G_4\right) = \sum_{x_i\in G_1}\sum_{x_j\in G_2}\sum_{x_k\in G_3} \sum_{x_l\in G_4} \frac{\mathbb{I}_{\left(i,j,k,l\right) \in \setQ} - \mathbb{I}_{\left(k,l,i,j\right) \in \setQ}}{\left|G_1\right|\left|G_2\right|\left|G_3\right|\left|G_4\right|} \,.
\end{align}

Based on our model for passive comparisons, where $\xi_{ijkl} \sim \text{Bernoulli}(p)$ is the indicator for observing tuple $(i,j,k,l)$, we may re-write the preference relation in~\eqref{app:eq:4--AL-pre} as  
\begin{align*}
\Wb_\setQ\left(G_1,G_2 \| G_3,G_4\right) = \sum_{x_i\in G_1}\sum_{x_j\in G_2}\sum_{x_k\in G_3} \sum_{x_l\in G_4} \frac{\xi_{ijkl}(\mathbb{I}_{(w_{ij}>w_{kl})} - \mathbb{I}_{(w_{ij}<w_{kl})})}{\left|G_1\right|\left|G_2\right|\left|G_3\right|\left|G_4\right|} \,.
\end{align*}

Subsequently, we use the above preference relation $\Wb_\setQ$ to define a similarity function $W$ in the following way.
Suppose that we have a disjoint partition $G_1,\ldots,G_K$ of $\setX$ and that we want to know which clusters should be merged next.
We define the similarity of clusters $G_p,G_q$, $p\neq q$, as
\begin{align}
\label{app:eq:4--AL}
W\left(G_p,G_q\right) = \sum_{\substack{r,s=1, r \neq s}}^K \frac{\Wb_\setQ\left(G_p,G_q \| G_r,G_s\right)}{K(K-1)}\,.
\end{align}
The underlying idea is that two clusters $G_p$ and $G_q$ are similar to each other if, on average, the pair is often preferred over the other possible cluster pairs.
The above similarity measure $W$, in conjunction with the hierarchical clustering principle (Algorithm~1 in the main paper), results in the proposed 4--AL algorithm.
Below, we restate and prove the exact recovery guarantee for 4--AL using passively obtained quadruplet comparisons.

\begin{reth}{\ref{th:exact4--AL}}[Exact recovery of planted hierarchy by 4--AL with passive comparisons\label{app:th:exact4--AL}] 
Let $\eta\in(0,1)$ and $\Delta = \frac{\delta}{2\sigma} e^{-L^2\delta^2/4\sigma^2}$. Assume the following: 
\\
(i) An initial step partitions $\setX$ into pure clusters of sizes in the range $[m,2m]$ for some $m \leq \frac12 N_0$.
\\
(ii) $\setQ$ is a passively obtained set of quadruplet comparisons, where each tuple $(i,j,k,l)$ is observed independently with probability $p > \displaystyle \frac{C}{m\Delta^2} \max\left\{\ln N, \frac1m \ln\left(\frac1\eta\right)\right\}$ for some constant $C>0$.  
\\Then,  with probability $1-\eta$, starting from the given initial partition and using $|\setQ| \leq pN^4$ number of passive comparisons, 4--AL exactly recovers the planted hierarchy.

In particular, if $L=\bigO{1}$, the above statement implies that, when $\frac\delta\sigma$ is a constant, 4--AL exactly recovers the planted hierarchy with probability $1-\eta$ using $\bigO{\frac{N^4\ln N}{m}}$ passive comparisons.
\end{reth}

\begin{proof}
The bound $|\setQ| < pN^4$ with probability $1-\frac\eta2$ is derived similarly to the bound on $|\setQ|$ in Theorem~\ref{app:th:exact4K--AL-passive}. Hence, we only prove the exact recovery guarantee.

We first analyze the algorithm under expectation. 
Assume that at some stage of the agglomerative iterations, we have a partition $G_1,\ldots,G_K$ of $\setX$.
Assume that the partition adheres to the ground truth, that is, either each $G_p$ is a subset of a pure cluster or an union of several pure clusters that corresponds to one of the nodes in the top $L$ levels of the true hierarchy.
Consider $p,q,r,s \in\{1,\ldots,K\}$ such that $p\neq q$, $r\neq s$, $\llca(G_p,G_q) = \ell$ and $\llca(G_r,G_s) = \ell'$.
From the definition of $\Wb_\setQ$, we have
\begin{align*}
\Eb[\Wb_\setQ\left(G_p,G_q \| G_r,G_s\right)] 
&= \sum_{x_i\in G_p} \sum_{x_j\in G_q} \sum_{x_k\in G_r} \sum_{x_l\in G_s}
\frac{p\big(2\Pb(w_{ij} > w_{kl}) - 1\big)}{ \left|G_p\right|\left|G_q\right|\left|G_r\right|\left|G_s\right|} 
\\&= \sum_{x_i\in G_p} \sum_{x_j\in G_q} \sum_{x_k\in G_r} \sum_{x_l\in G_s}
\frac{p\beta_{\ell-\ell'}}{ \left|G_p\right|\left|G_q\right|\left|G_r\right|\left|G_s\right|}  
\\&= p\beta_{\ell-\ell'} \,.
\end{align*}
Now, consider $p,q,p',q' \in\{1,\ldots,K\}$ such that $p\neq q$, $p'\neq q'$, $\llca(G_p,G_q) = \ell+1$ and $\llca(G_{p'},G_{q'}) = \ell$ for some $\ell\in\{0,1,\ldots,L-1\}$.
Thus, according to the planted model, one should merge $G_p,G_q$ before $G_{p'},G_{q'}$. 
We verify that this is indeed the case under expectation since
\begin{align*}
\Eb[W(G_p,G_q)]& - \Eb[W(G_{p'},G_{q'})] 
\\&= \frac{1}{K(K-1)} \sum_{\substack{r,s=1 \\ r \neq s}}^K \Eb[\Wb_\setQ\left(G_p,G_q \| G_r,G_s\right)] - \Eb[\Wb_\setQ\left(G_{p'},G_{q'} \| G_r,G_s\right)] \,.
\\&=  \frac{1}{K(K-1)} \sum_{\substack{r,s=1 \\ r \neq s}}^K p\beta_{\ell + 1 - \llca(G_r,G_s)} -  p\beta_{\ell - \llca(G_r,G_s)}
\\&> p\Delta,
\end{align*}
where the last step follows from arguments used in the proof of Theorem~\ref{app:th:exact4K--AL-passive}, which show that $\min\limits_{\ell\in[-L,L-1]} \beta_{\ell+1} - \beta_{\ell} > \Delta$, where $\beta_\ell$ is  defined in~\eqref{app:eq:pf-beta-def} and $\Delta$ is in the statement of the theorem.

Chaining of the above argument shows that $\Eb[W(G_p,G_q)] - \Eb[W(G_{p'},G_{q'})] >p\Delta$ whenever $\llca(G_p,G_q) > \llca(G_{p'},G_{q'})$.
Under the assumptions stated in Theorem~\ref{app:th:exact4--AL}, we later prove that with probability $1-\frac\eta2$,
\begin{align}
\big| W(G,G') - \Eb[W(G,G')] \big| \leq \frac{p\Delta}{2}
\label{app:eq:pf-4--AL1}
\end{align}
for every pair of clusters $G,G'$ formed during the agglomerative steps of the algorithm starting from the given pure clusters of size in the range $[m,2m]$.
Based on~\eqref{app:eq:pf-4--AL1} and the above argument, it is evident that $W(G_p,G_q) > W(G_{p'},G_{q'})$ whenever $\llca(G_p,G_q) > \llca(G_{p'},G_{q'})$ and, in particular, the cluster pair that achieves the maximum at any stage of iteration must be merged at the earliest according to the planted hierarchy.
This guarantees exact recovery of the planted hierarchy by the algorithm.

We now prove~\eqref{app:eq:pf-4--AL1}. For this, we first state a concentration inequality that we prove later. 
Let $G_1,G_2,G_3,G_4$ be four clusters, each of size in the range $[m,2m]$, such that $G_1,G_2$ are disjoint and so are $G_3,G_4$.
Then
\begin{align}
\label{app:eq:pf-4--AL2}
\Pb\left( \left| \Wb_\setQ(G_1,G_2 \| G_3, G_4) - \Eb[\Wb_\setQ(G_1,G_2 \| G_3, G_4)] \right| > \frac{p\Delta}{2} \right)
\leq 2\exp\left( 2\ln N -  \frac{p\Delta^2 m^2}{C'} \right)
\end{align}
for some absolute constant $C'>0$.
We wish to use~\eqref{app:eq:pf-4--AL2} to argue that with probability $1-\frac\eta2$, all clusters in the initial partition (assumed in the theorem) satisfy the condition $\left| \Wb_\setQ(G_1,G_2 \| G_3, G_4) - \Eb[\Wb_\setQ(G_1,G_2 \| G_3, G_4)] \right| \leq \frac{p\Delta}{2}$. 
Note that we do not know how the initial partition is achieved, but we can ensure that
\begin{align*}
\Pb&\bigg( \exists G_1,G_2,G_3,G_4: m \leq |G_1|,|G_2|,|G_3|,|G_4| \leq 2m, 
\\&\hskip20ex \left| \Wb_\setQ(G_1,G_2 \| G_3, G_4) - \Eb[\Wb_\setQ(G_1,G_2 \| G_3, G_4)] \right| > \frac{p\Delta}{2} \bigg)
\\&\leq \sum_{i_1,i_2,i_3,i_4 = m}^{2m} \binom{N}{i_1}\binom{N}{i_2}\binom{N}{i_3}\binom{N}{i_4} 2\exp\left( 2\ln N -  \frac{p\Delta^2 m^2}{C'} \right)
\\&\leq 2m^4 \left( \frac{eN}{m} \right)^{8m} \exp\left( 2\ln N -  \frac{p\Delta^2 m^2}{C'} \right).
\\&\leq C'' \exp\left(9m\ln N -  \frac{p\Delta^2 m^2}{C'} \right),
\end{align*}
where $C''>0$ is an absolute constant such that
$\sup\limits_{m\geq1} 2m^4 (\frac{e}{m})^{2m} < C''$.
The above probability is bounded by $\frac\eta2$ for $p > \displaystyle \frac{C}{m\Delta^2} \left(\ln N \vee \frac1m \ln\left(\frac1\eta\right)\right)$ for some constant $C>0$.
Thus, with probability $1-\frac\eta2$, we know that for every tuple of four clusters, obtained at initialization,  $\Wb_\setQ$ deviates from its mean by at most $\frac{p\Delta}{2}$.
In fact, the same deviation also holds when we merge some of these clusters.
For instance, let $G_1,G'_1,G_2,G_3,G_4$ be part of a partition at some stage and suppose $G_1,G'_1$ are merged. Then
\begin{align*}
\Wb_\setQ(G_1\cup G'_1, G_2 \| G_3,G_4) &= \frac{|G_1|}{|G_1|+|G'_1|} \Wb_\setQ(G_1, G_2 \| G_3,G_4) 
\\&\hskip20ex+ \frac{|G'_1|}{|G_1|+|G'_1|} \Wb_\setQ(G'_1, G_2 \| G_3,G_4),
\end{align*}
which is a convex combination of $\Wb_\setQ$ computed at the previous stage. Hence, if each of them deviates from its mean by at most $\frac{p\Delta}{2}$, then the convex combination after merging also deviates from its mean by at most $\frac{p\Delta}{2}$. 
The same also holds for other instances of merging throughout the hierarchy, which shows that with probability $1-\frac\eta2$, at any stage of agglomeration, we have $\left| \Wb_\setQ(G_p,G_q \| G_r, G_s) - \Eb[\Wb_\setQ(G_p,G_q \| G_r, G_s)] \right| < \frac{p\Delta}{2}$ for any tuple of four clusters in the partition.
Now, observe that $W(G_p,G_q)$ is an average of several $\Wb_\setQ$, and so,~\eqref{app:eq:pf-4--AL1} holds.

We complete the proof of Theorem~\ref{app:th:exact4--AL} by proving the concentration inequality in~\eqref{app:eq:pf-4--AL2}.
Since $w_{ij}=w_{kl}$ occurs with zero probability for any $i,j,k,l (i\neq j, k\neq l)$, we may write
\begin{align*}
&\left| \Wb_\setQ(G_1,G_2 \| G_3, G_4) - \Eb[\Wb_\setQ(G_1,G_2 \| G_3, G_4)] \right| 
\\& = \frac{2}{\left|G_1\right|\left|G_2\right|\left|G_3\right|\left|G_4\right|} \left| \sum_{x_i\in G_1} \sum_{x_j\in G_2} \sum_{x_k\in G_3} \sum_{x_l\in G_4}
 \left(\xi_{ijkl} \mathbb{I}_{(w_{ij}>w_{kl})} -  p\Pb(w_{ij} > w_{kl}) \right) \right| ,
\end{align*}
where $\xi_{ijkl}$ is the indicator of observing the comparison between $(i,j)$ and $(k,l)$.
Note that each term in the summation is a centred random variable in the range $[-1,1]$, and has variance bounded by $p$. 
Let us denote each of them by $B_{ijkl}$, and observe that they have dependencies among themselves.
We use the concentration technique  of \citet{janson2002infamous}.
Consider the dependency graph for these variables, which is a graph on $s=|G_1||G_2||G_3||G_4|$ vertices and two vertices are adjacent if they are dependent. 
Some of the vertices have degree $|G_1||G_2|-1$ (dependent with other variables with same $k,l$), while other vertices have degree $|G_3||G_4|-1$.
Let us denote the maximum degree by $d$.
One can find an equitable colouring for such a graph using $(d+1)$ colours, where equitable denotes that all colour classes are of nearly equal sizes $\lfloor \frac{s}{d+1}\rfloor$ or $\lceil \frac{s}{d+1}\rceil$. Denoting the colour classes by $\mathcal{C}_1,\ldots,\mathcal{C}_{d+1}$, we can bound the probability using the union bound and Bernstein's inequality as
\begin{align*}
\Pb&\left(\left| \Wb_\setQ(G_1,G_2 \| G_3, G_4) - \Eb[\Wb_\setQ(G_1,G_2 \| G_3, G_4)] \right|  > \frac{p\Delta}{2}\right) 
\\&= \Pb\left(\left| \sum_{i,j,k,l} B_{ijkl} \right| > \frac{sp\Delta}{4} \right)
\\&\leq \sum_{\ell=1}^{d+1}  \Pb\left(\left| \sum_{(i,j,k,l)\in\mathcal{C}_\ell} B_{ijkl} \right| > \frac{sp\Delta}{4(d+1)}\right)
\\&\leq \sum_{\ell=1}^{d+1} 2\exp\left(-\frac{\frac{s^2p^2\Delta^4}{16(d+1)^2}}{2p|\mathcal{C}_\ell| + \frac23 \frac{sp\Delta}{4(d+1)}}\right).
\end{align*}
The bound in~\eqref{app:eq:pf-4--AL2} follows by first noting that $|\mathcal{C}_\ell| \leq \frac{2s}{d+1}$, and then using the fact $\frac{s}{d+1}\geq \min\{|G_1| |G_2|, |G_3| |G_4|\} \geq m^2$. For the outer summation, we simply use $(d+1) \leq N^2$ to obtain the bound in~\eqref{app:eq:pf-4--AL2}.

To verify the claim for fixed $L$ and $\frac\delta\sigma$, we note that, in this case, $\Delta$ is constant and $N_0 = \bigOmega{N}$.
Hence, using $p= \frac{c\ln N}{m}$ for a large enough constant $c$ immediately leads to the exact recovery guarantee and the number of passive comparisons.
\end{proof}

\section{Details on the experiments}
\label{app:sec:moreresults}

In this section we present some details on the experiments that are not included in the main paper along with some additional plots and discussions.

\subsection{Planted Hierarchical Model}

\textbf{Evaluation function.} As a measure of performance we report the Averaged Adjusted Rand Index (AARI) between the ground truth hierarchy $\setC$ and the hierarchies $\setC^\prime$ learned by the different methods. Let $\setC^\ell$ and ${\setC^\prime}^\ell$ be the partitions of $\setX$ at level $\ell$ of the hierarchies, then:
\begin{align*}
\text{AARI}\left(\setC,\setC^\prime\right) = \frac{1}{L} \sum_{\ell \in \left\lbrace 1,\ldots,L \right\rbrace} \text{ARI}\left(\setC^\ell,{\setC^\prime}^\ell\right)
\end{align*}
where ARI is the Adjusted Rand Index \citep{hubert1985comparing}, a widely used measure to compare partitions.
We use the average across the different levels $\setC^\ell$ and ${\setC^\prime}^\ell$ to take into account the hierarchical structure.
The AARI takes values in the interval $\left[0,1\right]$ and the higher the value the more similar the hierarchies are. $\text{AARI}\left(\setC,\setC^\prime\right) = 1$ implies that the two hierarchies are identical.
For all the experiments we report the mean and the standard deviation of $10$ repetitions.

\begin{figure}
    \centering
    \subfloat[$p = 0.01$\label{app:fig:init1}]{\includegraphics[height=3.35cm]{{Images/th_30_1}.png}}
    \subfloat[$p = 0.02$\label{app:fig:init2}]{\includegraphics[height=3.35cm]{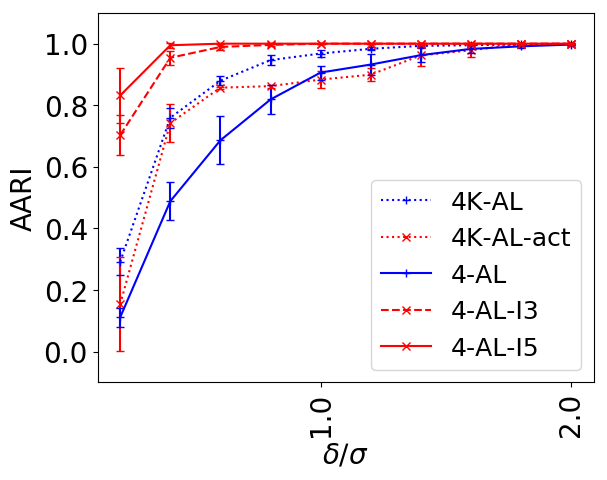}}
    \subfloat[$p = 0.03$\label{app:fig:init3}]{\includegraphics[height=3.35cm]{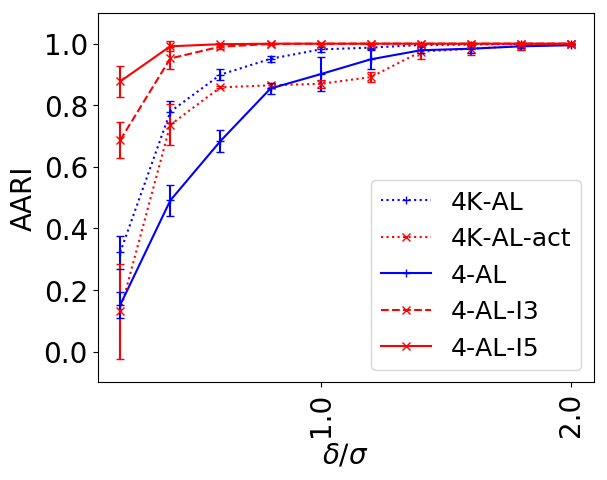}}\\
    \subfloat[$p = 0.04$\label{app:fig:init4}]{\includegraphics[height=3.35cm]{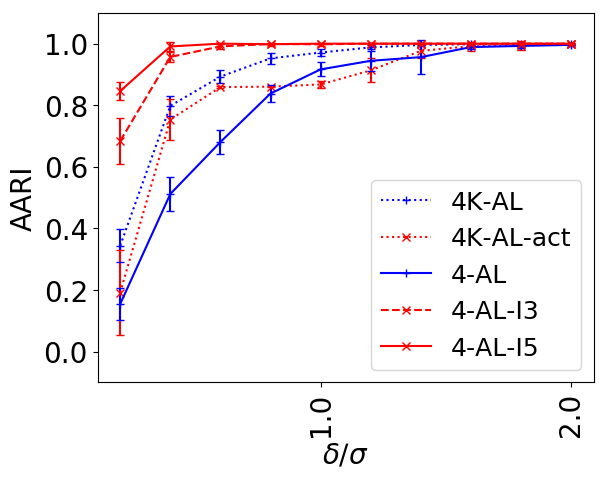}}
    \subfloat[$p = 0.05$\label{app:fig:init5}]{\includegraphics[height=3.35cm]{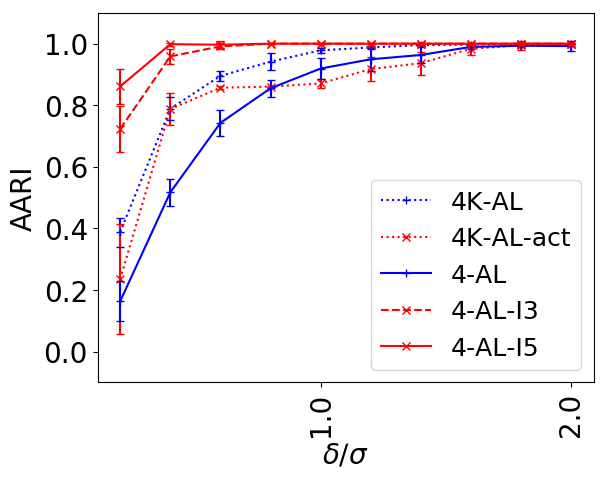}}
    \subfloat[$p = 0.06$\label{app:fig:init6}]{\includegraphics[height=3.35cm]{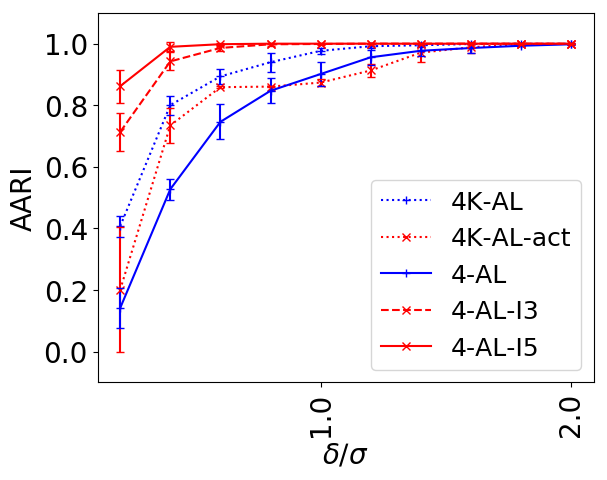}}\\
    \subfloat[$p = 0.07$\label{app:fig:init7}]{\includegraphics[height=3.35cm]{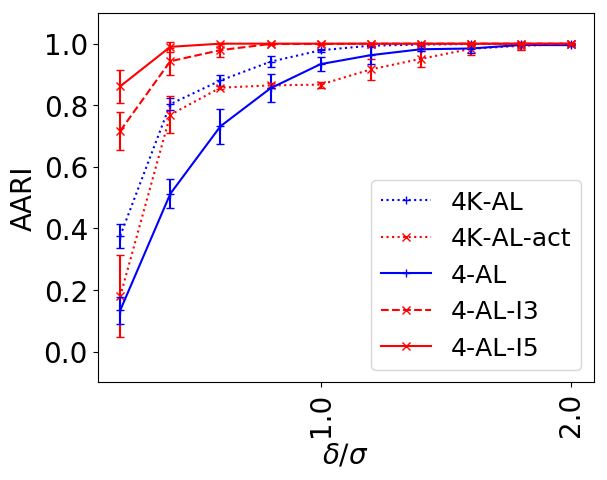}}
    \subfloat[$p = 0.08$\label{app:fig:init8}]{\includegraphics[height=3.35cm]{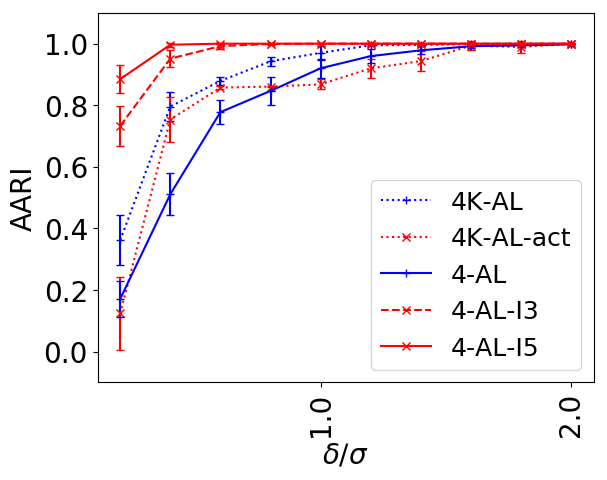}}
    \subfloat[$p = 0.09$\label{app:fig:init9}]{\includegraphics[height=3.35cm]{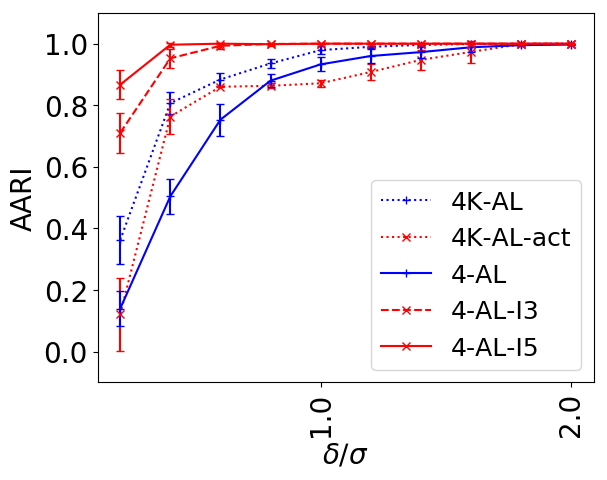}}\\
    \subfloat[$p = 0.1$\label{app:fig:init10}]{\includegraphics[height=3.35cm]{{Images/th_30_10}.png}}
    \subfloat[$p = 1$\label{app:fig:init100}]{\includegraphics[height=3.35cm]{{Images/th_30_100}.png}}\\
	\caption{AARI of the proposed methods (higher is better) on data obtained from the planted hierarchical model with $\mu = 0.8$, $\sigma = 0.1$, $L = 3$, $N_0 = 30$ and different sampling proportions $p$. Best viewed in color.
\label{app:fig:theoreticalexperimentsAARI}}
\end{figure}

\textbf{Results.} In Figure~\ref{app:fig:theoreticalexperimentsAARI} we present supplementary results for the planted hierarchical model, that is with $p \in \left\{ 0.01,0.02,\ldots,0.1,1 \right\}$.
Firstly, similar to the theory, SL can hardly recover the planted hierarchy, even for large values of $\frac{\delta}{\sigma}$. CL performs better than SL, which is not evident from the theory. This suggests that a better sufficient condition might be possible for CL.
We observe that 4K--AL, 4K--AL--act, and, 4--AL are able to exactly recover the true hierarchy for smaller signal-to-noise ratio and their performances do not degrade much when the number of sampled comparisons is reduced.
Finally, as expected, the best methods are 4--AL--I3 and 4--AL--I5. They use large initial clusters but recover the true hierarchy even for very small values of $\frac{\delta}{\sigma}$.

\subsection{Standard Clustering Datasets}

\textbf{Data.} We provide some details on the datasets used in the paper. We evaluate the different approaches on $3$ different datasets commonly used in hierarchical clustering: Zoo, Glass and 20news \citep{heller2005bayesian,vikram2016interactive}.
The Zoo dataset is composed of $100$ animals with $16$ features (it originally contains $101$ animals but we chose to remove the `girl' entry since we feel that it does not fit in a Zoo dataset).
The Glass dataset has $9$ features for $214$ examples.
The 20news dataset is composed of $11314$ news articles.
Following \citet{vikram2016interactive} we pre-processed the 20news dataset using a bag of words approach followed by PCA to retain $100$ relevant features.
We randomly sampled $200$ examples for hierarchical clustering.
To fit the comparison-based setting we generate the quadruplet comparisons using the cosine similarity:
\begin{align*}
w_{ij} = \frac{\left\langle \vx_i,\vx_j\right\rangle}{\left\|\vx_i\right\|\left\|\vx_i\right\|}
\end{align*}
where $\vx_i$ and $\vx_j$ are the representations of objects $x_i$ and $x_j$ and $\left\langle\cdot,\cdot\right\rangle$ is the dot product.
Since it is not realistic to assume that all the comparisons are available, we use the procedure described in Section~\ref{subsec:cbframework} in the main paper to passively obtain a proportion $p \in \left\lbrace 0.01, 0.02, \ldots, 0.1\right\rbrace$ of all the quadruplets.
Note that tSTE-AL and FORTE-AL are based on ordinal embedding methods that use triplet comparisons of the form ``object $i$ is more similar to object $j$ than to object $k$'', that is $w_{ij} > w_{ik}$, rather than quadruplet comparisons.
Nevertheless, we can use the same procedure than for the quadruplets to generate the same proportion of triplets that we can use in tSTE and FORTE.
To the best of our knowledge, there does not exist ordinal embedding methods based only on quadruplet comparisons.

\textbf{Evaluation function.}
Contrary to the planted hierarchical model we do not have access to a ground-truth hierarchy and thus we cannot use the AARI measure to evaluate the performance of the methods.
Instead we use the recently proposed Dasgupta's cost \citep{dasgupta2016cost} that has been specifically designed to evaluate hierarchical clustering methods.
Given a base similarity measure $w$, the cost of a hierarchy $\setC$ is
\begin{align*}
\text{cost}(\setC,w) = \sum_{x_,x_j \in \setX} w_{ij} \left| \setC^{lca}(x_i,x_j) \right|
\end{align*}
where $w_{ij}$ is the similarity between $x_i$ and $x_j$ and $\setC^{lca}(x_i,x_j)$ is the smallest cluster containing both $x_i$ and $x_j$ in the hierarchy.
The idea of this cost is that similar objects that are merged higher in the hierarchy should be penalized.
Hence, a lower cost indicates a better hierarchy.
A low cost is achieved if similar objects (high $w_{ij}$) are merged towards the bottom of the tree (small $\setC^{lca}(x_i,x_j)$), and vice-versa.
Hence, a lower value of the cost indicates a better hierarchy.
For all the experiments we report the mean and the standard deviation of $10$ repetitions.

\begin{figure}
    \centering
    \subfloat[$\text{Dimension} = 2$\label{app:fig:Zoodim2}]{\includegraphics[height=4cm]{{Images/zoo_2}.png}}
    \subfloat[$\text{Dimension} = 3$\label{app:fig:Zoodim3}]{\includegraphics[height=4cm]{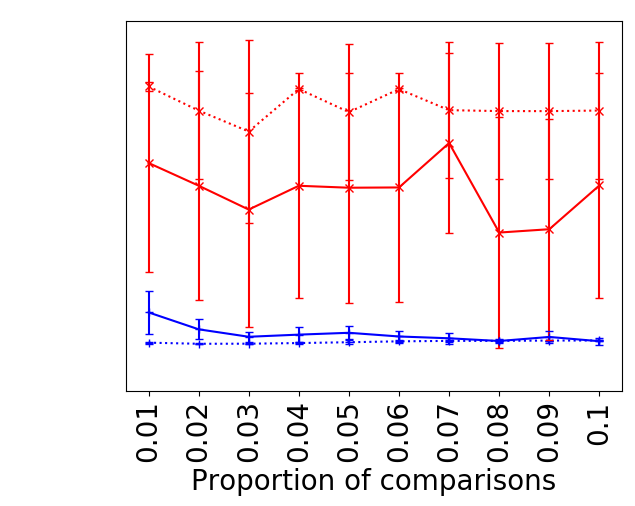}} \\
    \subfloat[$\text{Dimension} = 4$\label{app:fig:Zoodim4}]{\includegraphics[height=4cm]{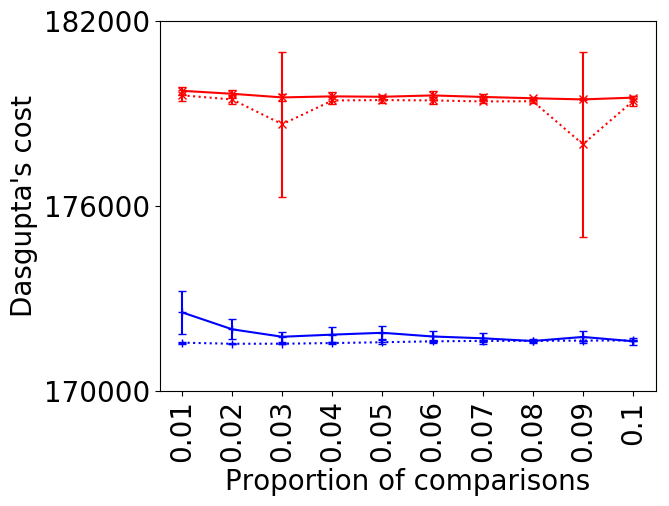}}
    \subfloat[$\text{Dimension} = 5$\label{app:fig:Zoodim5}]{\includegraphics[height=4cm]{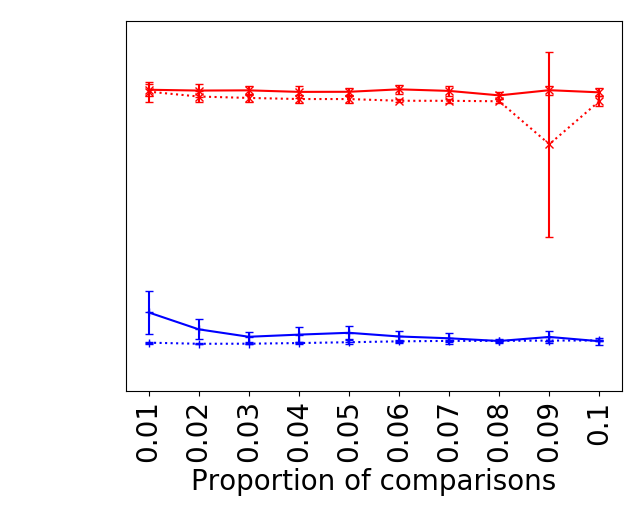}}
	\caption{Dasgupta's score of the different methods on the Zoo dataset with increasing embedding dimensions for FORTE--AL and tSTE--AL.  Best viewed in color.\label{app:fig:Zoodim}}
\end{figure}

\begin{figure}
    \centering
    \subfloat[$\text{Dimension} = 2$\label{app:fig:Glassdim2}]{\includegraphics[height=4cm]{{Images/glass_2}.png}}
    \subfloat[$\text{Dimension} = 3$\label{app:fig:Glassdim3}]{\includegraphics[height=4cm]{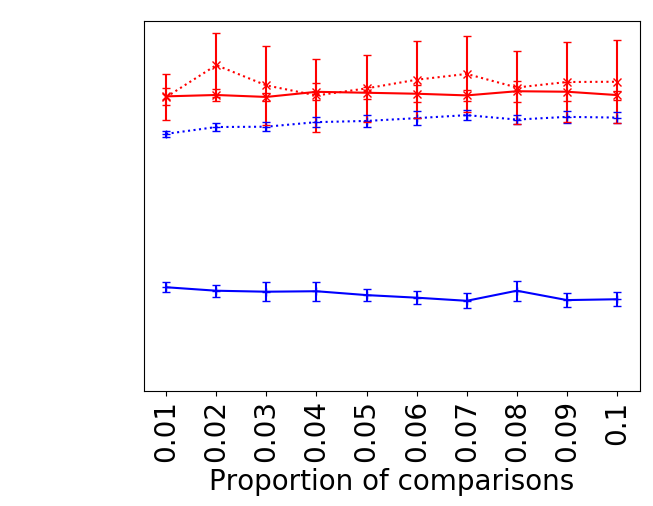}}\\
    \subfloat[$\text{Dimension} = 4$\label{app:fig:Glassdim4}]{\includegraphics[height=4cm]{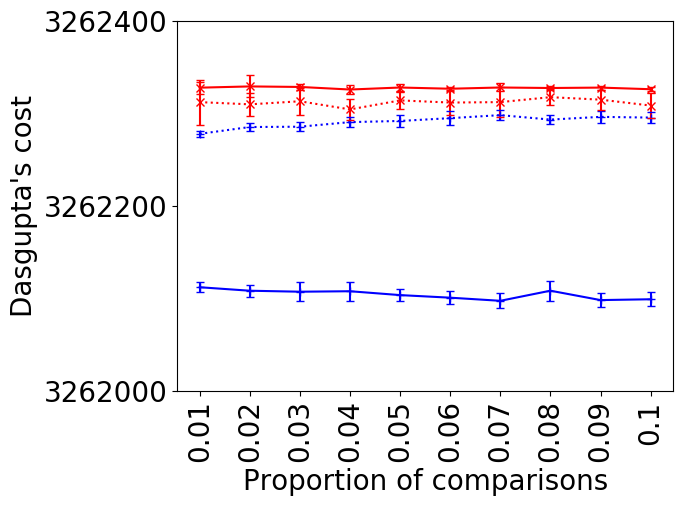}}
    \subfloat[$\text{Dimension} = 5$\label{app:fig:Glassdim5}]{\includegraphics[height=4cm]{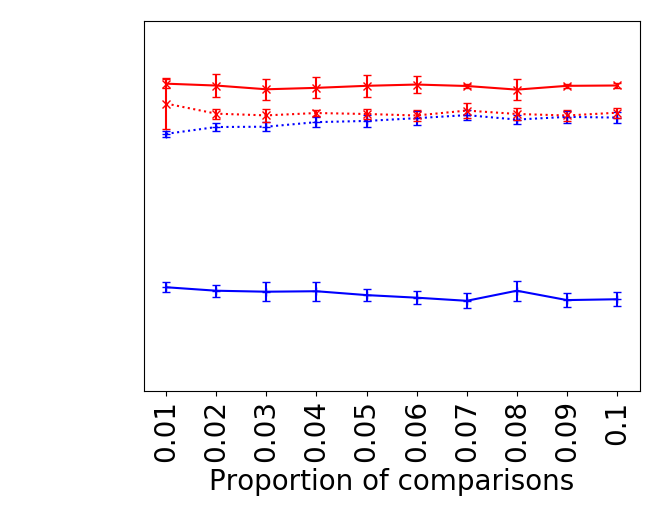}}
	\caption{Dasgupta's score of the different methods on the Glass dataset with increasing embedding dimensions for FORTE--AL and tSTE--AL. Best viewed in color.\label{app:fig:Glassdim}}
\end{figure}

\begin{figure}
    \centering
    \subfloat[$\text{Dimension} = 2$\label{app:fig:20newsdim2}]{\includegraphics[height=4cm]{{Images/20news200_2}.png}}
    \subfloat[$\text{Dimension} = 3$\label{app:fig:20newsdim3}]{\includegraphics[height=4cm]{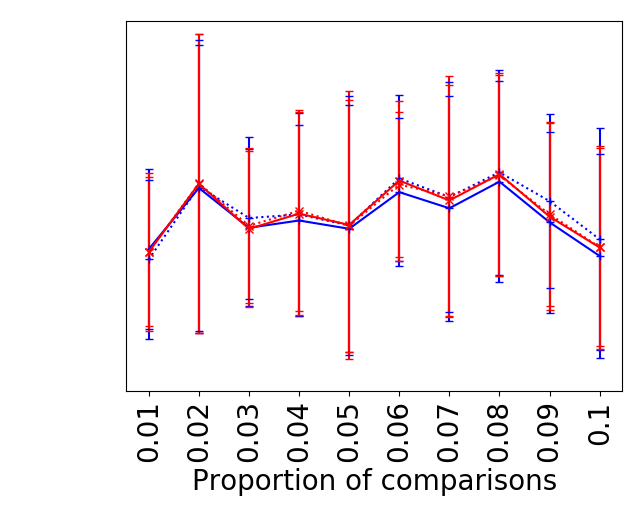}}\\
    \subfloat[$\text{Dimension} = 4$\label{app:fig:20newsdim4}]{\includegraphics[height=4cm]{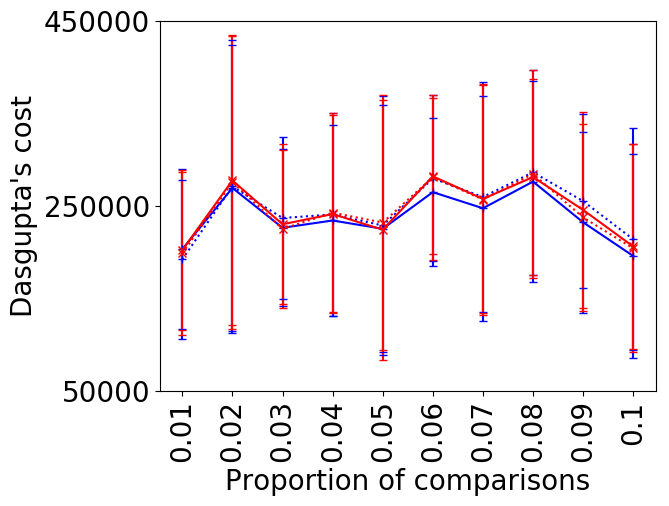}}
    \subfloat[$\text{Dimension} = 5$\label{app:fig:20newsdim5}]{\includegraphics[height=4cm]{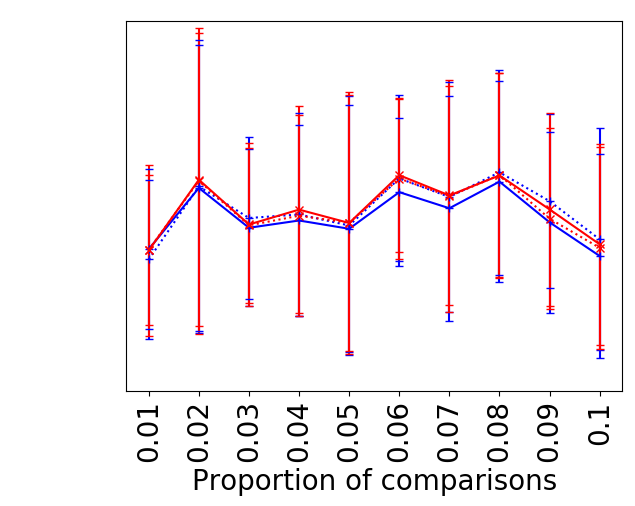}}
	\caption{Dasgupta's score of the different methods on the 20news dataset with increasing embedding dimensions for FORTE--AL and tSTE--AL. Best viewed in color.\label{app:fig:20newsdim}}
\end{figure}

\textbf{Results.} In Figures~\ref{app:fig:Zoodim},~\ref{app:fig:Glassdim},~and~\ref{app:fig:20newsdim} we present supplementary results for the standard clustering datasets.
We note that the proportion of comparisons does not have a large impact as the results are, on average, stable across all regimes.
Our methods are either comparable or better than the embedding-based ones.
Our methods do not need to first embed the examples and thus do not impose a strong Euclidean structure on the data.
The impact of this structure is more or less pronounced depending on the dataset.
Furthermore, the performance of tSTE-AL and FORTE-AL depends on the embedding dimension that should be carefully chosen. For example, on Zoo, the performance of tSTE drops with increasing dimension. Similarly, on Glass, FORTE seems to perform slightly better for larger dimensions. Unfortunately, in clustering, tuning parameters can be difficult as there is no ground-truth.

\subsection{Comparison-based datasets}

The Car dataset \citep{kleindessner2017lens} is composed of $60$ different type of cars and $6056$ ordinal comparisons, collected via crowd-sourcing, of the form \textit{Which car is most central among the three $x_i$, $x_j$ and $x_k$?}.
These statements translate easily to the triplet setting: if $x_i$ is most central in the set of three then we recover two triplets $(j,i,k)$ and $(k,i,j)$.
Then triplet comparisons further translate into quadruplet comparisons by noticing that the triplet $(i,j,k)$ corresponds to the quadruplet $(i,j,i,k)$.
Overall we obtained $12112$ comparisons that we used to learn a hierarchy among the cars.
The hierarchies obtained by 4K--AL, 4--AL, FORTE--AL and tSTE--AL are attached to this supplementary as png files. The names of the files are respectively cars.4K--AL.png, cars.4--AL.png, cars.FORTE--AL.embedding\_dimension.png and cars.tSTE--AL.embedding\_dimension.png.

\end{document}